\documentclass[journal,article,submit,moreauthors,pdftex]{article} 
\usepackage{geometry}
\usepackage{placeins}
\usepackage{balance}
\usepackage{mystyle}
\usepackage{tikz}
\usepackage{xcolor}
\usepackage{authblk}
\usepackage{amsmath}
\usepackage{booktabs}
\usepackage{multirow}
\usepackage[outline]{contour}
\usepackage[hyphens]{url}
\usepackage{hyperref}
\usepackage[backend=bibtex, maxbibnames=99]{biblatex} 
\usepackage{subcaption}

\newcommand\ARXIV[2]{#1}

\title{Crop Rotation Modeling for Deep Learning-Based \\ Parcel Classification from Satellite Time Series}

\author{Félix Quinton, Loic Landrieu}

\affil{%
LASTIG, Univ. Gustave Eiffel, ENSG, IGN, F-94160 Saint-Mandé, France
}

\date{}

\begin{document}
\maketitle
\begin{abstract}
While annual crop rotations play a crucial role for agricultural optimization, they have been largely ignored for automated crop type mapping .
In this paper, we take advantage of the increasing quantity of annotated satellite data to propose the first deep learning approach modeling simultaneously the inter- and intra-annual agricultural dynamics of parcel classification. Along with simple training adjustments, our model provides an improvement of over $6.6$ mIoU points over the current state-of-the-art of crop classification. Furthermore, we release the first large-scale multi-year agricultural dataset with over $300\,000$ annotated parcels.
\end{abstract}

\section{Introduction}

The Common Agricultural Policy (CAP) is responsible for allocating agricultural subsidies in the European Union, which nears $50$ billion euros each year \cite{CAP}. Consequently, monitoring the crop types for subsidy allocation represents a significant challenge for payment agencies, which have encouraged the development of automated crop classification tools based on machine learning \cite{loudjani2001artificial}. 
In particular, The Sentinels for Common Agricultural Policy (Sen4CAP) project \cite{koetz2019sen4cap} aims to provide EU member states with algorithmic solutions and best practice studies on crop monitoring based on satellite data from the Sentinel constellation \cite{drusch2012sentinel}.
Despite the inherent difficulty of differentiating between the complex growth patterns of plants, this task is made possible by the nearly limitless access to data and annotations. 
Indeed, Sentinel-2 offers multi-spectral observations at a high revisit time of five days {on average}, which are particularly appropriate for characterizing the complex spectral and temporal characteristics of crop phenology. Moreover, farmers declare the crop cultivated in each of their parcels every year. This represents over $10$ million of annotations each year for France alone \cite{RPG}, all openly accessible in the Land-Parcel Identification System (LPIS).
However, the sheer scale of the problem raises interesting computational  challenges: Sentinel-2 gathers over $25$Tb of data each year over Europe.

\begin{figure}[ht!]
 \begin{tabular}{ccc}
 \begin{tikzpicture}
 \node[anchor=south west,inner sep=0] (image) at (0,0) {\includegraphics[width=\ARXIV{0.32}{0.23}\textwidth]{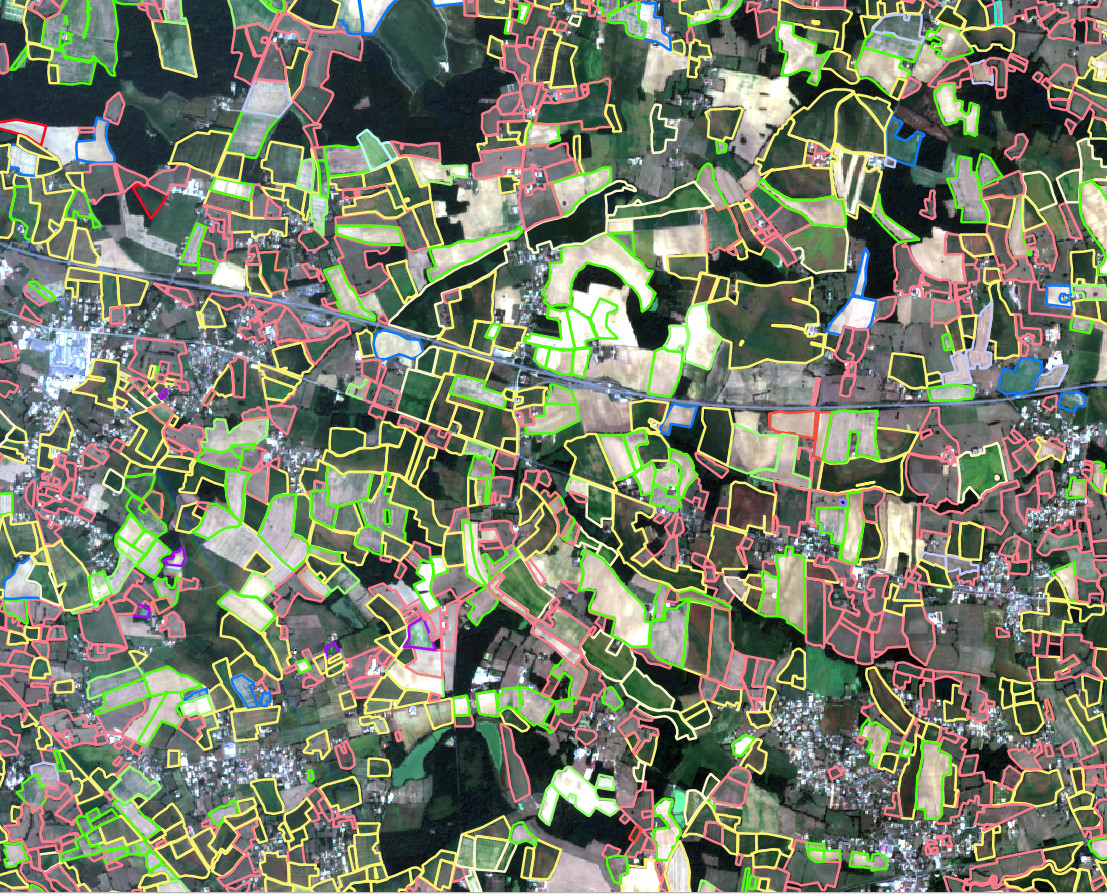}};
        \begin{scope}[x={(image.south east)},y={(image.north west)}]
        \node[fill=white, draw=none, text=black] (n1) at (0.2,0.9) {\Large \bf 2018} ;
        \end{scope}
 \end{tikzpicture}
 \begin{tikzpicture}
 \node[anchor=south west,inner sep=0] (image) at (0,0) {\includegraphics[width=\ARXIV{0.32}{0.23}\textwidth]{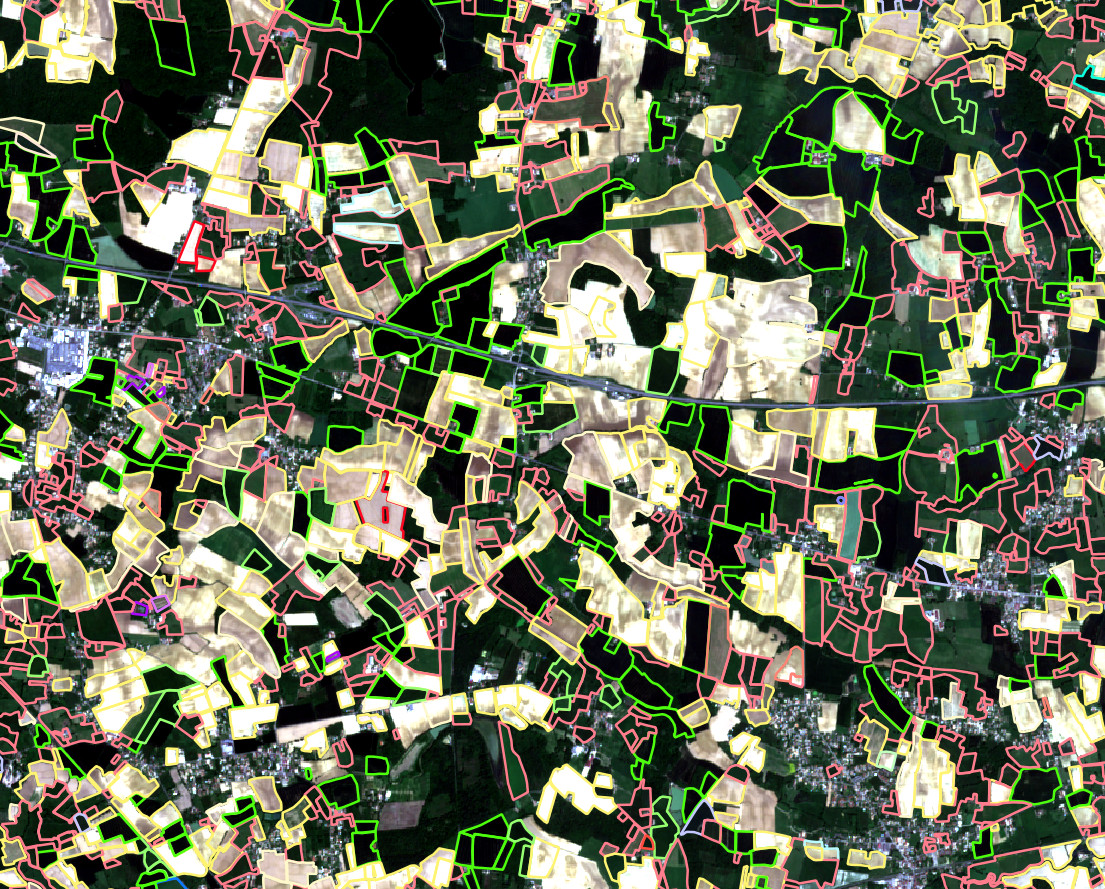}};
        \begin{scope}[x={(image.south east)},y={(image.north west)}]
        \node[fill=white, draw=none, text=black] (n1) at (0.2,0.9) {\Large \bf 2019} ;
        \end{scope}
 \end{tikzpicture}
 \begin{tikzpicture}
 \node[anchor=south west,inner sep=0] (image) at (0,0) {\includegraphics[width=\ARXIV{0.32}{0.23}\textwidth]{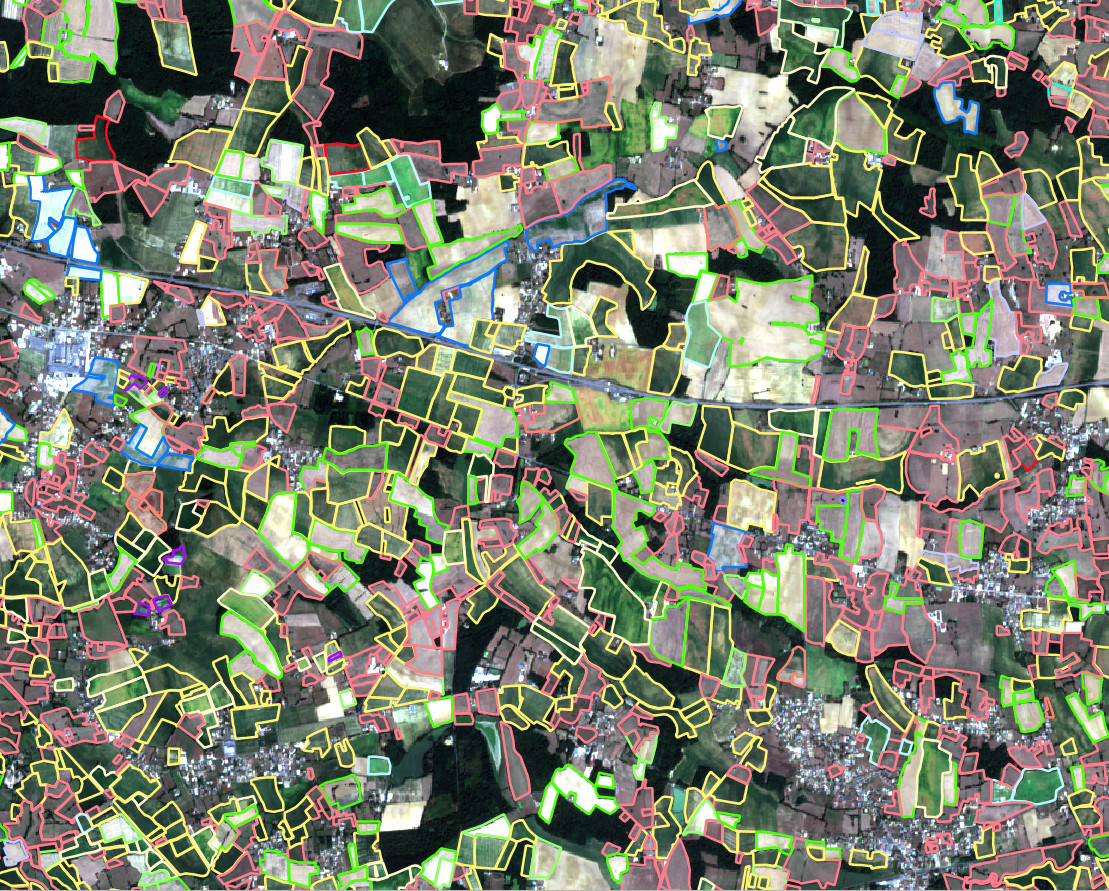}};
        \begin{scope}[x={(image.south east)},y={(image.north west)}]
        \node[fill=white, draw=none, text=black] (n1) at (0.2,0.9) {\Large \bf 2020} ;
        \node[fill=none, draw=none, text=black] (n1) at (0.9,0.9) {\includegraphics[width=\ARXIV{0.03}{0.03}\textwidth]{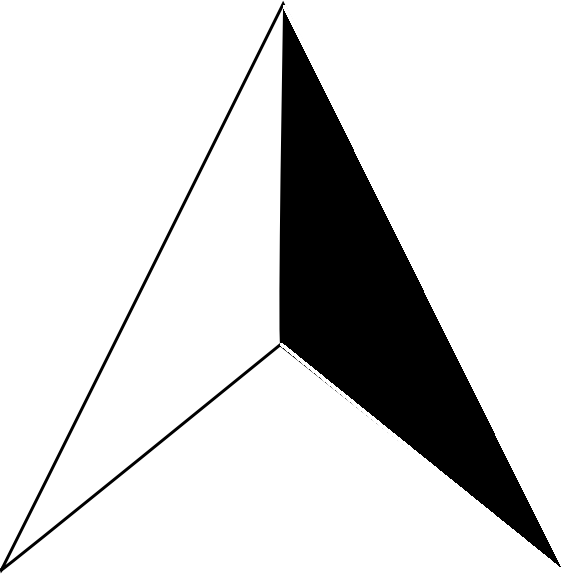}} ;
        \node[fill=none, draw=none, text=black] (n1) at (0.9,0.80) {\contour{white}{N}} ;
        \draw[fill=white, text=black, draw=none] (0.75,0.05) rectangle (0.95,0.13);
        \draw[<->, fill=white, text=black, draw=black] (0.75,0.10) -- (0.95,0.10);
        \draw[-, draw=black] (0.85,0.09) -- (0.85,0.11);
        \node[fill=none, text=black, draw=none] at (0.85,0.07)  {\tiny 1 km};
        \end{scope}
 \end{tikzpicture}
 &
 \\
 \multicolumn{3}{c}{
 \begin{tabular}{rlrlrlrl}
 \definecolor{tempcolor}{rgb}{0.98,0.29,0.23}
           \tikz \fill[fill=tempcolor, scale=0.3, draw=black] (0,0) rectangle (1,1);
           & \small{Oats Win} 
           &
           \definecolor{tempcolor}{rgb}{1,0.95,0.05}
           \tikz \fill[fill=tempcolor, scale=0.3, draw=black] (0,0) rectangle (1,1); 
           & \small{Oats Sum}
           &
           \definecolor{tempcolor}{rgb}{0.3,1,0.05}
           \tikz \fill[fill=tempcolor, scale=0.3, draw=black] (0,0) rectangle (1,1); 
           & \small{Wheats}
           &
           \definecolor{tempcolor}{rgb}{0.05,1,0.87}
           \tikz \fill[fill=tempcolor, scale=0.3, draw=black] (0,0) rectangle (1,1);
           & \small{Mixed cereal}
           \\
           \definecolor{tempcolor}{rgb}{0.05,0.47,1}
           \tikz \fill[fill=tempcolor, scale=0.3, draw=black] (0,0) rectangle (1,1);
           & \small{Rapeseed} 
           &
           \definecolor{tempcolor}{rgb}{0.66,0.04,1}
           \tikz \fill[fill=tempcolor, scale=0.3, draw=black] (0,0) rectangle (1,1); 
           & \small{Flow/Frui/Vege}
           &
           \definecolor{tempcolor}{rgb}{1,0.03,0.13}
           \tikz \fill[fill=tempcolor, scale=0.3, draw=black] (0,0) rectangle (1,1); 
           & \small{Leguminous}
           &
           \definecolor{tempcolor}{rgb}{1,0.52,0.45}
           \tikz \fill[fill=tempcolor, scale=0.3, draw=black] (0,0) rectangle (1,1);
           & \small{Alfalfa}
           \\
           \definecolor{tempcolor}{rgb}{1,0.96,0.45}
           \tikz \fill[fill=tempcolor, scale=0.3, draw=black] (0,0) rectangle (1,1);
           & \small{Maize} 
           &
           \definecolor{tempcolor}{rgb}{1,0.81,0.87}
           \tikz \fill[fill=tempcolor, scale=0.3, draw=black] (0,0) rectangle (1,1); 
           & \small{Barley Win}
           &
           \definecolor{tempcolor}{rgb}{0.47,1,0.96}
           \tikz \fill[fill=tempcolor, scale=0.3, draw=black] (0,0) rectangle (1,1); 
           & \small{Barley Sum}
           &
           \definecolor{tempcolor}{rgb}{1,0.55,1}
           \tikz \fill[fill=tempcolor, scale=0.3, draw=black] (0,0) rectangle (1,1);
           & \small{Wood Pastures}
           \\
           \definecolor{tempcolor}{rgb}{0.87,0.46,1}
           \tikz \fill[fill=tempcolor, scale=0.3, draw=black] (0,0) rectangle (1,1);
           & \small{Potato} 
           &
           \definecolor{tempcolor}{rgb}{1,0.49,0.55}
           \tikz \fill[fill=tempcolor, scale=0.3, draw=black] (0,0) rectangle (1,1); 
           & \small{Meadow}
           &
           \definecolor{tempcolor}{rgb}{1,0.91,0.85}
           \tikz \fill[fill=tempcolor, scale=0.3, draw=black] (0,0) rectangle (1,1); 
           & \small{Rye}
           &
           \definecolor{tempcolor}{rgb}{1,0.97,0.79}
           \tikz \fill[fill=tempcolor, scale=0.3, draw=black] (0,0) rectangle (1,1);
           & \small{Soybean}
           \\
           \definecolor{tempcolor}{rgb}{0.75,1,0.74}
           \tikz \fill[fill=tempcolor, scale=0.3, draw=black] (0,0) rectangle (1,1);
           & \small{Sorghum} 
           &
           \definecolor{tempcolor}{rgb}{0.73,0.98,1}
           \tikz \fill[fill=tempcolor, scale=0.3, draw=black] (0,0) rectangle (1,1); 
           & \small{Sunflower}
           &
           \definecolor{tempcolor}{rgb}{0.76,0.79,1}
           \tikz \fill[fill=tempcolor, scale=0.3, draw=black] (0,0) rectangle (1,1); 
           & \small{Triticale}
           &
           \definecolor{tempcolor}{rgb}{1,0.81,0.87}
           \tikz \fill[fill=tempcolor, scale=0.3, draw=black] (0,0) rectangle (1,1);
           & \small{Vineyard}
      \end{tabular}
      }
 \end{tabular}
\caption{{\bf Multi-Year Sentinel-2 Data.}
Details of our area of interest for the three years studied in this article. {The crop type of each parcel is represented by} the color of a polygon following their contour according to the legend above. 
This color code is used throughout this article for all figures representing cultivated crops.
}
\label{fig:teaser}
\end{figure}
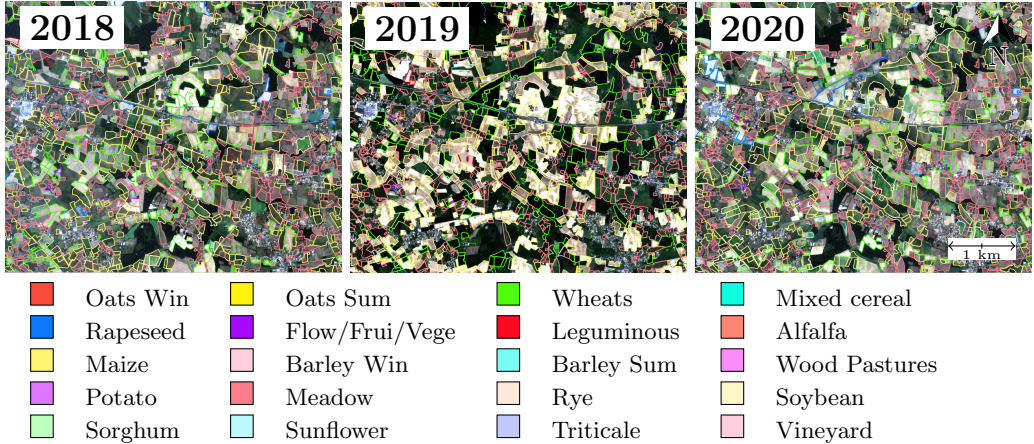

The state-of-the-art of {yearly} parcel-based crop type classification from Satellite Image Time Series (SITS) is particularly dynamic, especially since the adoption of deep learning methods \cite{garnot2020satellite,pelletier2019deep,russwurm2020self}. However, most methods operate on a single year worth of {temporal acquisitions} and ignore inter-annual crop rotations. In this paper, we propose the first deep learning framework for classifying {yearly} crop types from {information spanning several years, as represented in \figref{fig:teaser}}. We show that we can improve their predictions by a large margin with straightforward alterations of the top-performing models and their training protocols.\\

\paragraph{ Single-Year Crop-Type Classification.}  
Single-year crop-type classification involves the classification of the crop grown in a parcel from a single year worth of observation.
Pre-deep learning  parcel-based classification methods rely on such as support vector machines \cite{zheng2015support} or random forests \cite{vuolo2018much} operating on handcrafted descriptors such as the Normalized Difference Vegetation Index. The temporal dynamics are typically handled with stacking \cite{vuolo2018much}, probabilistic graphical models \cite{siachalou2015hidden}, or dynamic warping method \cite{belgiu2018sentinel}.

In conjunction with growing data availability, the adoption of deep learning-based methods has allowed for a large increase in performance for parcel-based crop classification. The spatial dimension of parcels is typically handled with convolutional neural networks \cite{kussul2017deep}, parcel-based statistics \cite{russwurm2020self}, or set-based encoders \cite{garnot2019time}. The temporal dynamics are modeled with temporal convolutions \cite{pelletier2019deep}, recurrent neural networks \cite{garnot2019time}, hybrid convolutional-recurrent networks \cite{russwurm2018multi}, and temporal attention \cite{russwurm2020self, garnot2020satellite, yuan2020self}.

Multiple recent studies \cite{kondmann2021denethor, garnot2020lightweight, garnot2020satellite, schneider2020re, garnot2021panoptic} have solidified the PSE+LTAE (Pixel Set Encoder \& Lightweight Temporal Attention) as the state-of-the-art of crop type classification for these reasons. Furthermore, this network is particularly parsimonious in terms of computation and memory usage, which proves well suited for training on multi-year data. Finally, the code is available (\url{https://github.com/VSainteuf/lightweight-temporal-attention-pytorch}, accessed on 10/10/21). For these reasons, we choose to use this network as the basis for our analysis and design modifications.\\

\paragraph{ Multi-Year Agricultural Optimization.}
Most of the literature on multi-year crop rotation focuses on agricultural optimization, \ie improving agricultural practices aiming to improve yields. These models generate suggested rotations  according to expert knowledge \cite{dury2012models}, handcrafted rules \cite{dogliotti2003rotat}, or statistical analysis \cite{myersmodeling}, while other models are based on a physical analysis of the soil composition \cite{brankatschk2015modeling} such as the nitrogen cycle \cite{detlefsen2007modelling}. Aurbacher and Dabbert also take a simple economic model into account in their study \cite{aurbacher2011generating}. 
More sophisticated models combine different sources of knowledge for better suggestions, such as ROTOR \cite{bachinger2007rotor} or CropRota \cite{Schonhar2011croprota}. The RPG Explorer software \cite{levavasseur2016rpg} uses a second-order Markov Chain for a more advanced statistical analysis of rotations. 

Given the popularity of these tools, it is clear that the careful choice of cultivated crops can substantially impact agricultural yields and is the object of meticulous attention from farmers. This is reinforced by the multi-model, multi-country meta-study of Kollas \etal \cite{kollas2015crop}, showing that multi-year modeling allows for a large increase in yield prediction. Consequently, we posit that a classification model with access to multi-year data will learn inter-annual patterns to improve its accuracy.\\

\paragraph{ Multi-Year Crop Type Classification.}
Multi-year crop type classification refers to leveraging information (satellite observations, past declarations) to improve the classification of the grown crop type in agricultural parcels.
Osman \etal \cite{osman2015assessment} propose to use probabilistic Markov models to predict the most probable crop type from the sequence of past cultivated crops of the previous 3 to 5 years. 
 Giordano \etal \cite{giordano2018temporal} and Bailly \etal \cite{bailly2018crop} propose to model the multi-year rotation with a second-order chain-Conditional Random Field (CRF).
 Finally, Yaramasu \etal \cite{yaramasu2020pre} are the first to propose to analyze multi-year data with a deep convolutional-recurrent model.
 However, they only choose one image per year and hence do not model both inter- and intra-annual dynamics.
 {In contrast, we propose to explicitly our model operates at both the intra-annual scale by using the sequence of yearly observation and the inter-annual scale by considering past declarations}. 
 
We list here the main contributions of this paper:
\begin{itemize}
    \item We propose a straightforward training scheme to leverage multi-year data and show its impact on {yearly} agricultural parcel classification.
    \item We introduce a modified attention-based temporal encoder able to model both inter- and intra-annual dynamics of agricultural parcels, yielding a large improvement in precision. 
    \item We present the first open-access multi-year dataset \cite{zenodo} for crop classification based on Sentinel-2 images, along with the full implementation of our model.
    \item Our code is open-source and accessible at the following repository:
    \url{https://github.com/felixquinton1/deep-crop-rotation}.
\end{itemize}
i
\section{Materials and Methods}

We present our dataset and proposed method to model multi-year SITS, along with several baseline methods to assess the performance of its components.
We denote by $[1, I]$ the set of years for which satellite observations are available to us and use the compact \emph{pixel-set} format to represent the SITS. For a given parcel and a year $i \in [1, I]$, we denote the corresponding SITS by a tensor $x^i$ of size $C \times S \times T_i$, with $C$ the number of spectral channels, $S$ the number of pixels within the parcel, and $T_i$ the number of temporal observation available for the year $i$. Likewise, we denote by $l^i \in \{0,1\}^L$ the one-hot-encoded label at year $i$, denoting which kind of crop is cultivated in the considered parcel among a set $L$ of crop types. {Note that, in this article, we focus on the prediction of the main culture, \ie only one crop type per year.}

\subsection{Dataset}
\begin{figure}[h]
    \centering
    \begin{tabular}{cc}
    \begin{subfigure}{\ARXIV{0.45}{0.35}\textwidth}
    
        \centering
        
        \begin{tikzpicture}
 \node[anchor=south west,inner sep=0] (image) at (0,0) {\includegraphics[width=\textwidth]{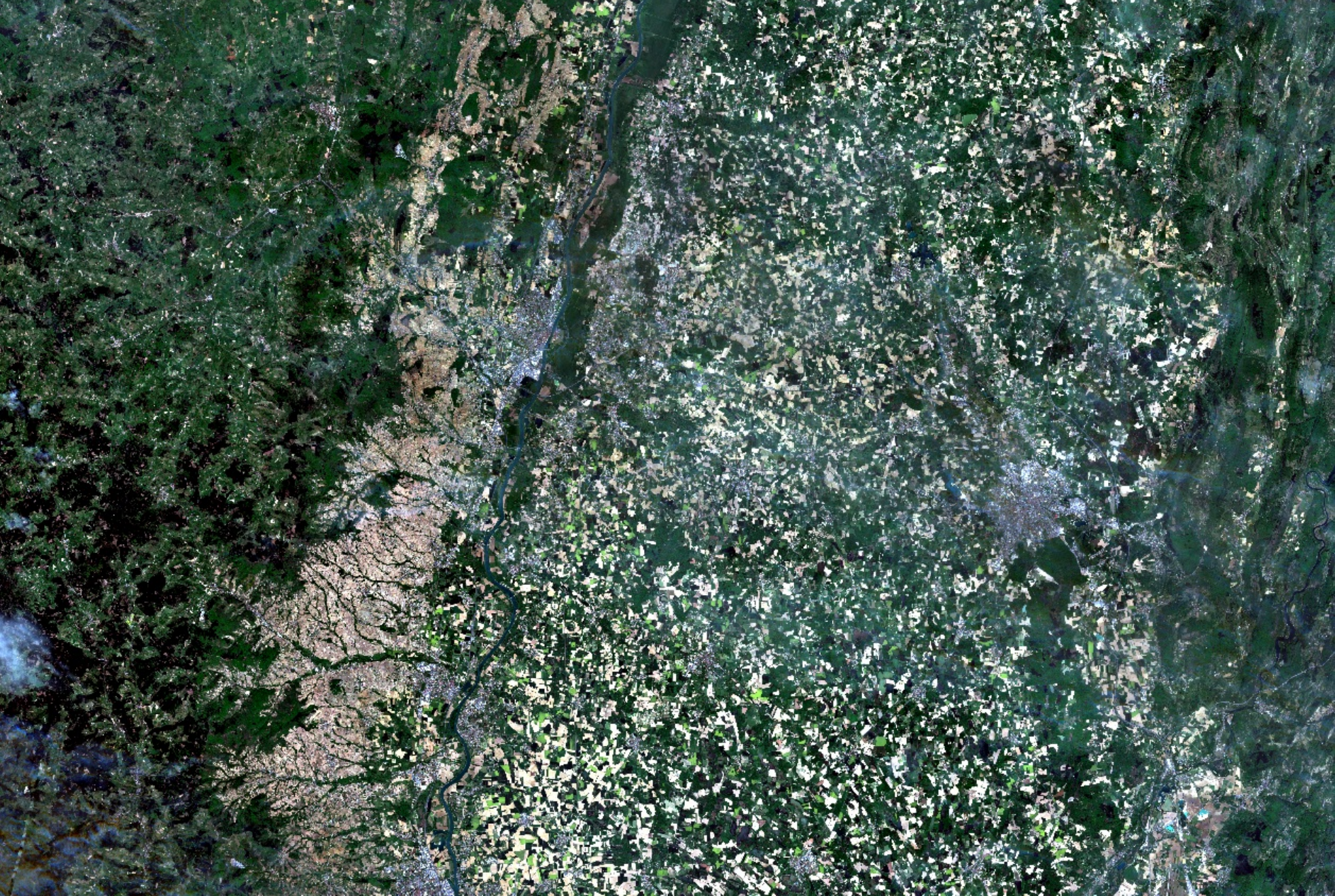}};
    \begin{scope}[x={(image.south east)},y={(image.north west)}]
        \draw[fill=none, draw=red] (0.5,0.65) rectangle (0.55,0.7);
        \node[fill=none, draw=none, text=black] (n1) at (0.9,0.9) {\includegraphics[width=\ARXIV{0.1}{0.1}\textwidth]{images/north.png}} ;
        \node[fill=none, draw=none, text=black] (n1) at (0.9,0.80) {\contour{white}{N}} ;
        \draw[fill=white, text=black, draw=none] (0.75,0.05) rectangle (0.95,0.13);
        \draw[<->, fill=white, text=black, draw=black] (0.75,0.10) -- (0.95,0.10);
        \draw[-, draw=black] (0.85,0.09) -- (0.85,0.11);
        \node[fill=none, text=black, draw=none] at (0.85,0.07)  {\tiny 20 km};
        \end{scope}
 \end{tikzpicture}
 \\
    \end{subfigure}
         & 
    \begin{subfigure}{\ARXIV{0.47}{0.35}\textwidth}
    \begin{tikzpicture}
    \node[anchor=south west,inner sep=0] (image) at (0,0) {\includegraphics[width=\textwidth]{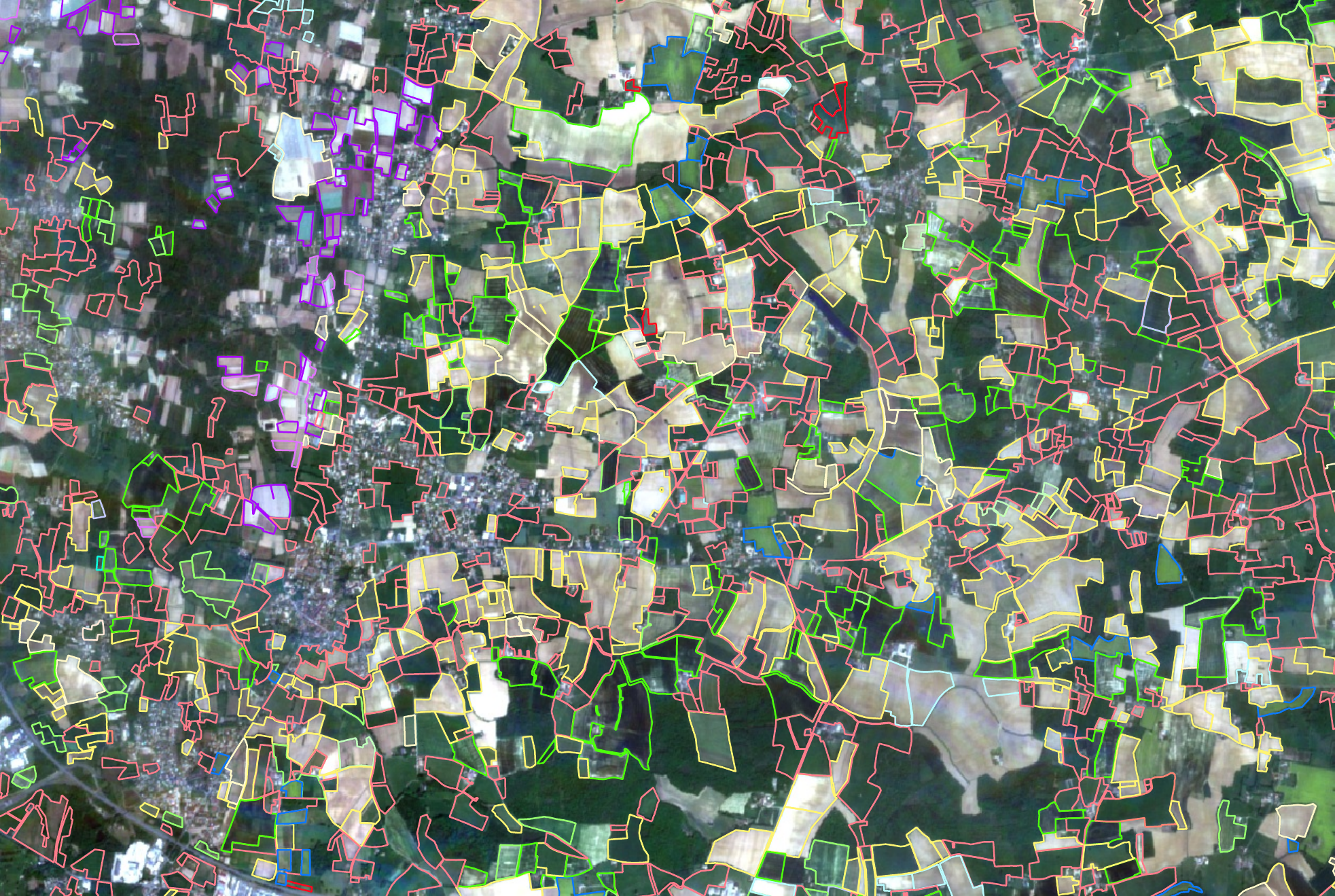}};
    \begin{scope}[x={(image.south east)},y={(image.north west)}]
        \draw[fill=none, draw=red] (0.5,0.65) rectangle (0.55,0.7);
        \node[fill=none, draw=none, text=black] (n1) at (0.9,0.9) {\includegraphics[width=\ARXIV{0.1}{0.1}\textwidth]{images/north.png}} ;
        \node[fill=none, draw=none, text=black] (n1) at (0.9,0.80) {\contour{white}{N}} ;
        \draw[fill=white, text=black, draw=none] (0.75,0.05) rectangle (0.95,0.13);
        \draw[<->, fill=white, text=black, draw=black] (0.75,0.10) -- (0.95,0.10);
        \draw[-, draw=black] (0.85,0.09) -- (0.85,0.11);
        \node[fill=none, text=black, draw=none] at (0.85,0.07)  {\tiny 1 km};
    \end{scope}
    \end{tikzpicture}
    \end{subfigure} \\
    \begin{subfigure}{\ARXIV{0.47}{0.2}\textwidth}
    \caption{}
    \label{fig:view:large}
    \end{subfigure}
    &
    \begin{subfigure}{\ARXIV{0.45}{0.2}\textwidth}
    \caption{}
    \label{fig:view:zoom}
    \end{subfigure}
     \\
    \end{tabular}
    \caption{\textbf{Area of Interest.} The studied parcels are taken from the 31TFM Sentinel-2 tile, covering  $110\times 110$ km and containing over $103\,602$ parcels meeting our size, shape, and stability criteria.
    {(\textbf{a})} Large view of the tile. (\textbf{b}) Detail of the area.
    }
    \label{fig:view}
\end{figure}

Our proposed dataset, represented in \figref{fig:view}, is based on parcels within the 31TFM Sentinel-2 tile, covering an area of $110 \times 110$  $\text{km}^{2}$ in the South East of France {(centered around $4.31$N, $46.44$E in WGS84). This area is in the Auvergne-Rhône-Alpes region, a major producer of cereal with over $54\,000$ha of corn and $30\,000$ha of wheat. Extensive livestock production makes meadow the most common crop type with over $60\%$ of declared parcels in the LPIS.  The most frequent crop rotations are permanent cultures (meadows, vineyards, pasture) and alternating between corn, wheat, and rapeseed.}

\begin{figure}[ht!]
    \centering
    \begin{tabular}{ccc}
    \begin{subfigure}{\ARXIV{0.31}{0.2}\textwidth}
        \centering\includegraphics[width=1\textwidth]{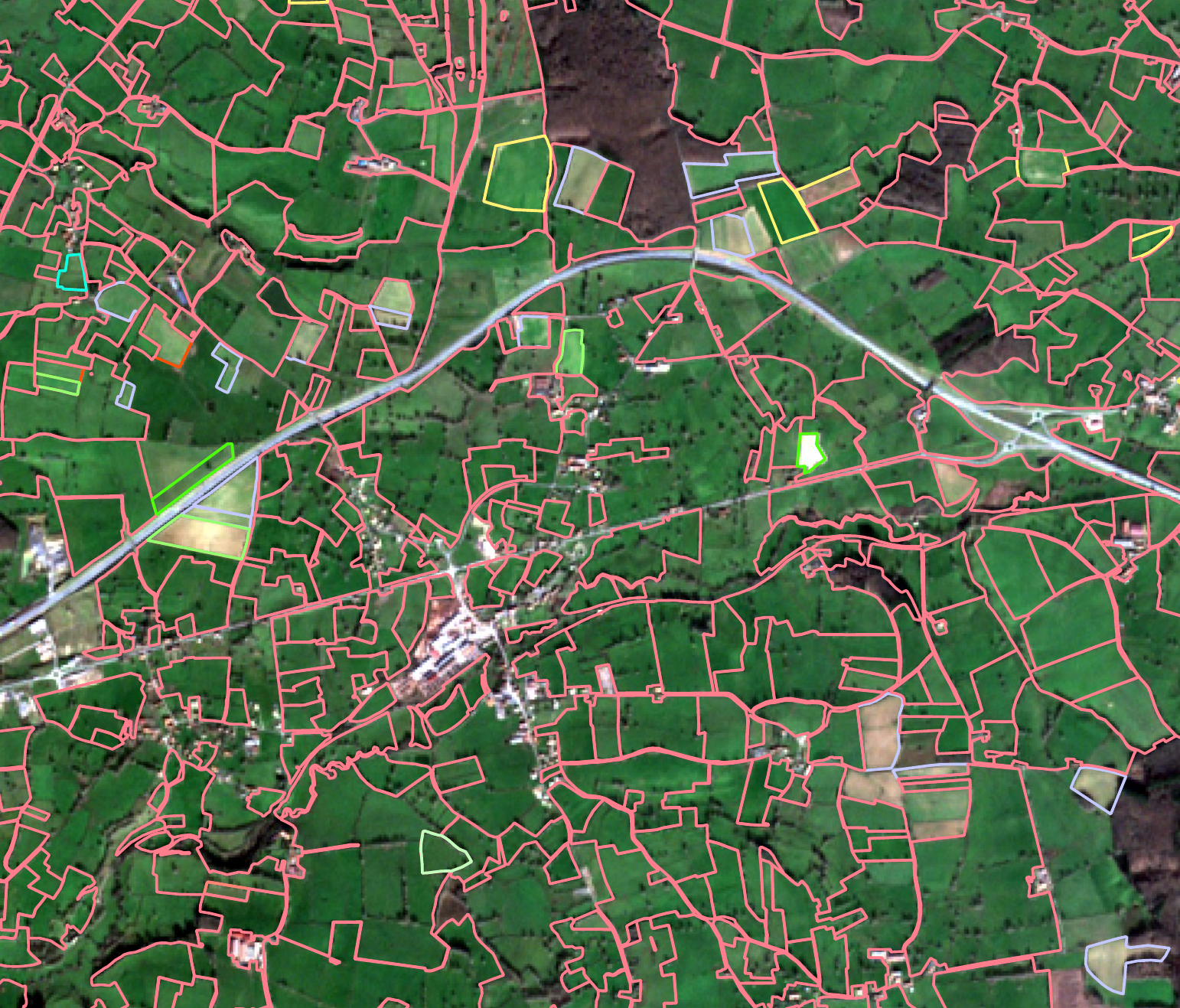}
    \end{subfigure}
         & 
    \begin{subfigure}{\ARXIV{0.31}{0.2}\textwidth}
        \centering\includegraphics[width=1\textwidth]{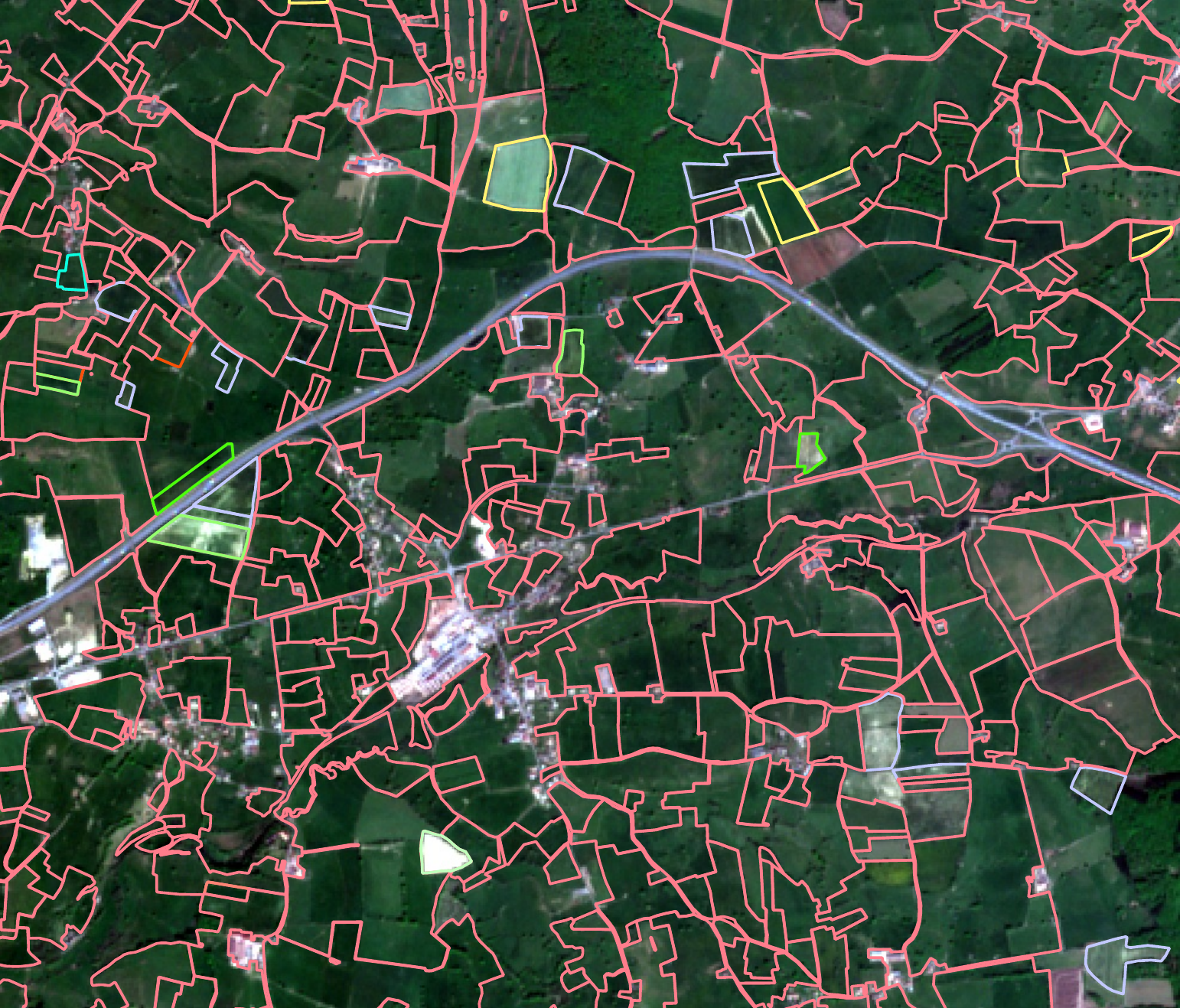}
    \end{subfigure}
    & 
    \begin{subfigure}{\ARXIV{0.31}{0.2}\textwidth}
        \centering
        \begin{tikzpicture}
    \node[anchor=south west,inner sep=0] (image) at (0,0) {\includegraphics[width=\textwidth]{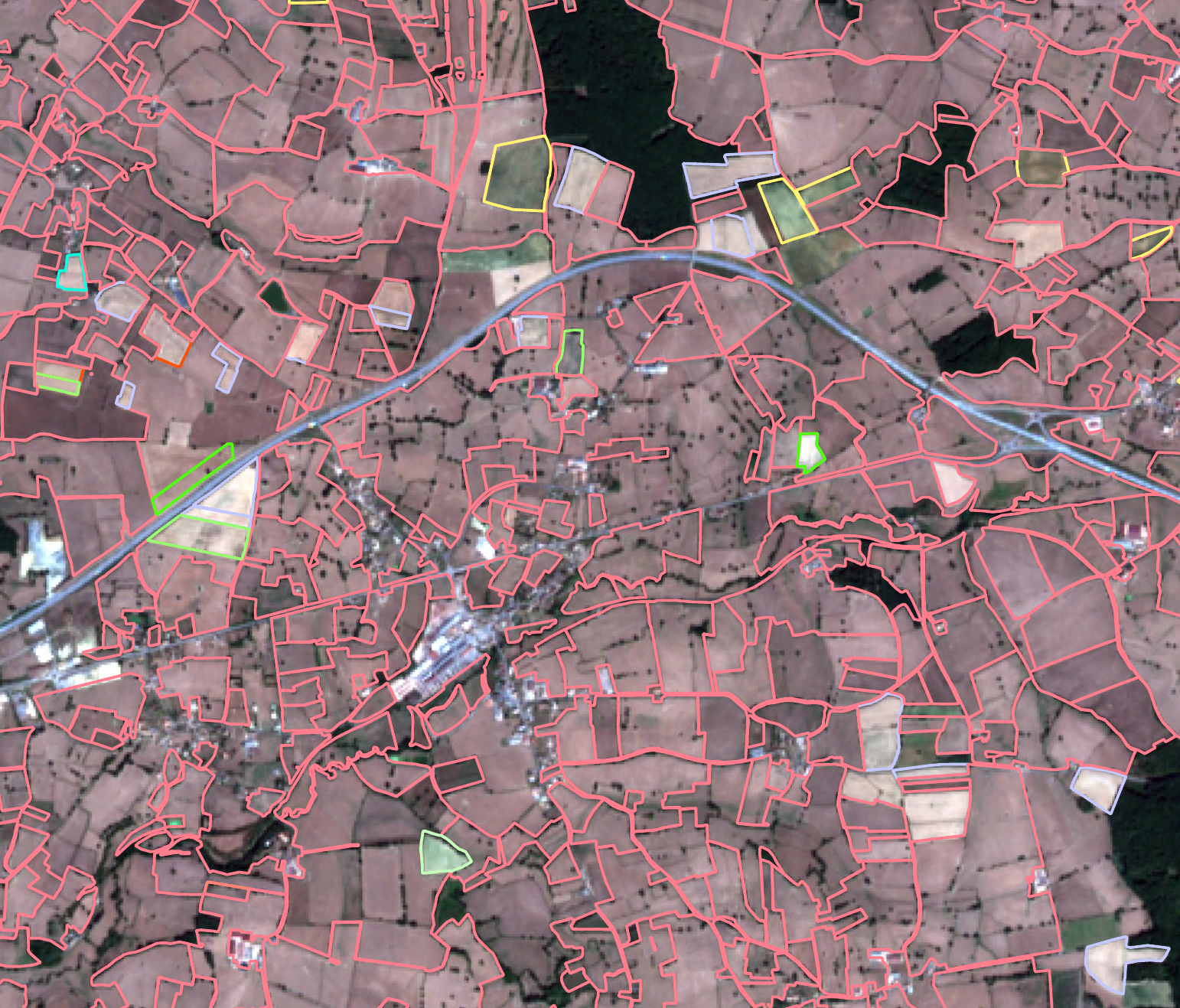}};
    \begin{scope}[x={(image.south east)},y={(image.north west)}]
        \draw[fill=none, draw=red] (0.5,0.65) rectangle (0.55,0.7);
        \node[fill=none, draw=none, text=black] (n1) at (0.9,0.9) {\includegraphics[width=\ARXIV{0.1}{0.1}\textwidth]{images/north.png}} ;
        \node[fill=none, draw=none, text=black] (n1) at (0.9,0.80) {\contour{white}{N}} ;
        \draw[fill=white, text=black, draw=none] (0.65,0.05) rectangle (0.95,0.13);
        \draw[<->, fill=white, text=black, draw=black] (0.65,0.10) -- (0.95,0.10);
        \draw[-, draw=black] (0.8,0.09) -- (0.8,0.11);
        \node[fill=none, text=black, draw=none] at (0.8,0.07)  {\tiny 1 km};
    \end{scope}
    \end{tikzpicture}
    \end{subfigure}\\ \\
    
    \begin{subfigure}{\ARXIV{0.31}{0.2}\textwidth}
        \centering\includegraphics[width=1\textwidth]{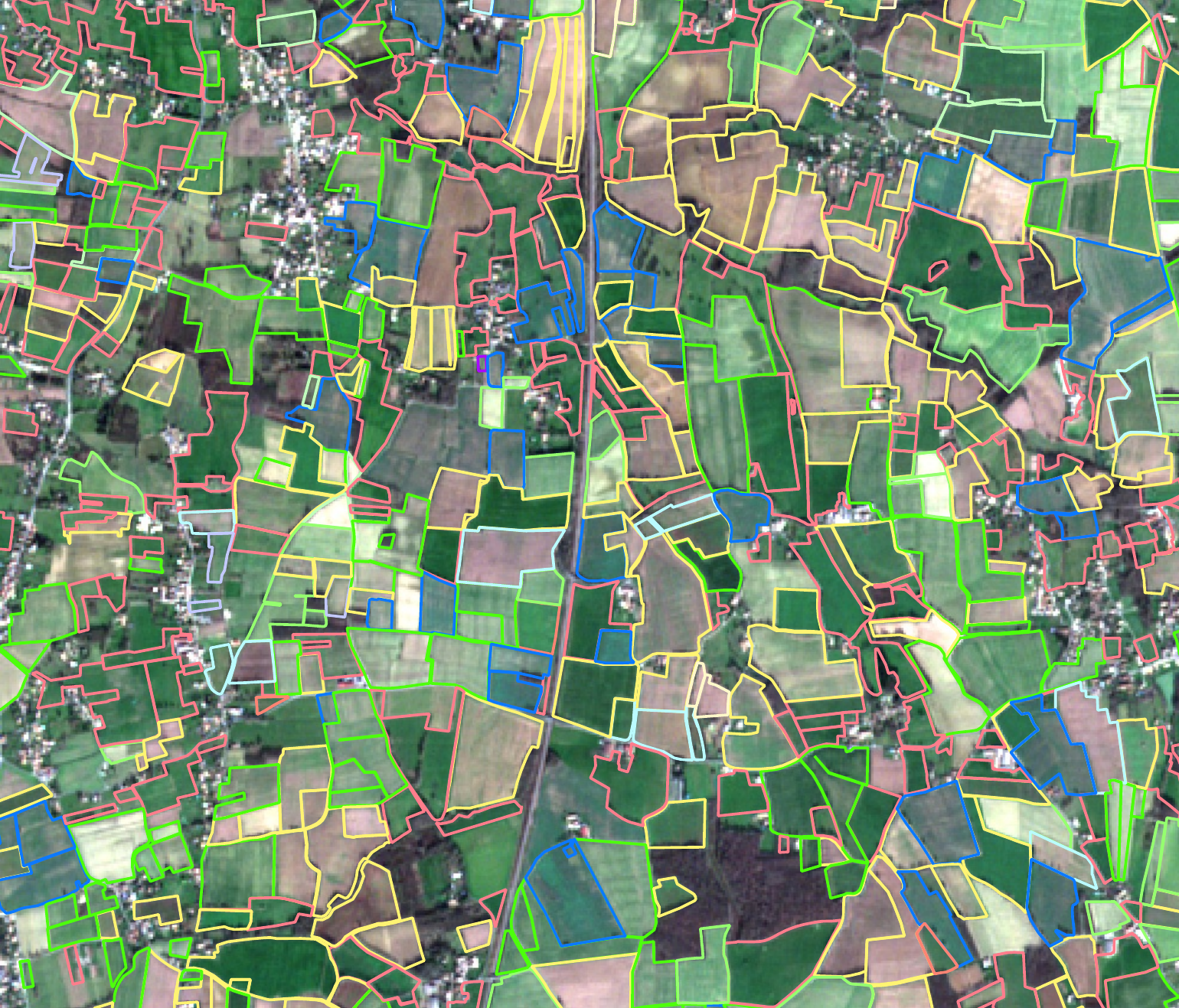}
    \end{subfigure}
         & 
    \begin{subfigure}{\ARXIV{0.31}{0.2}\textwidth}
        \centering\includegraphics[width=1\textwidth]{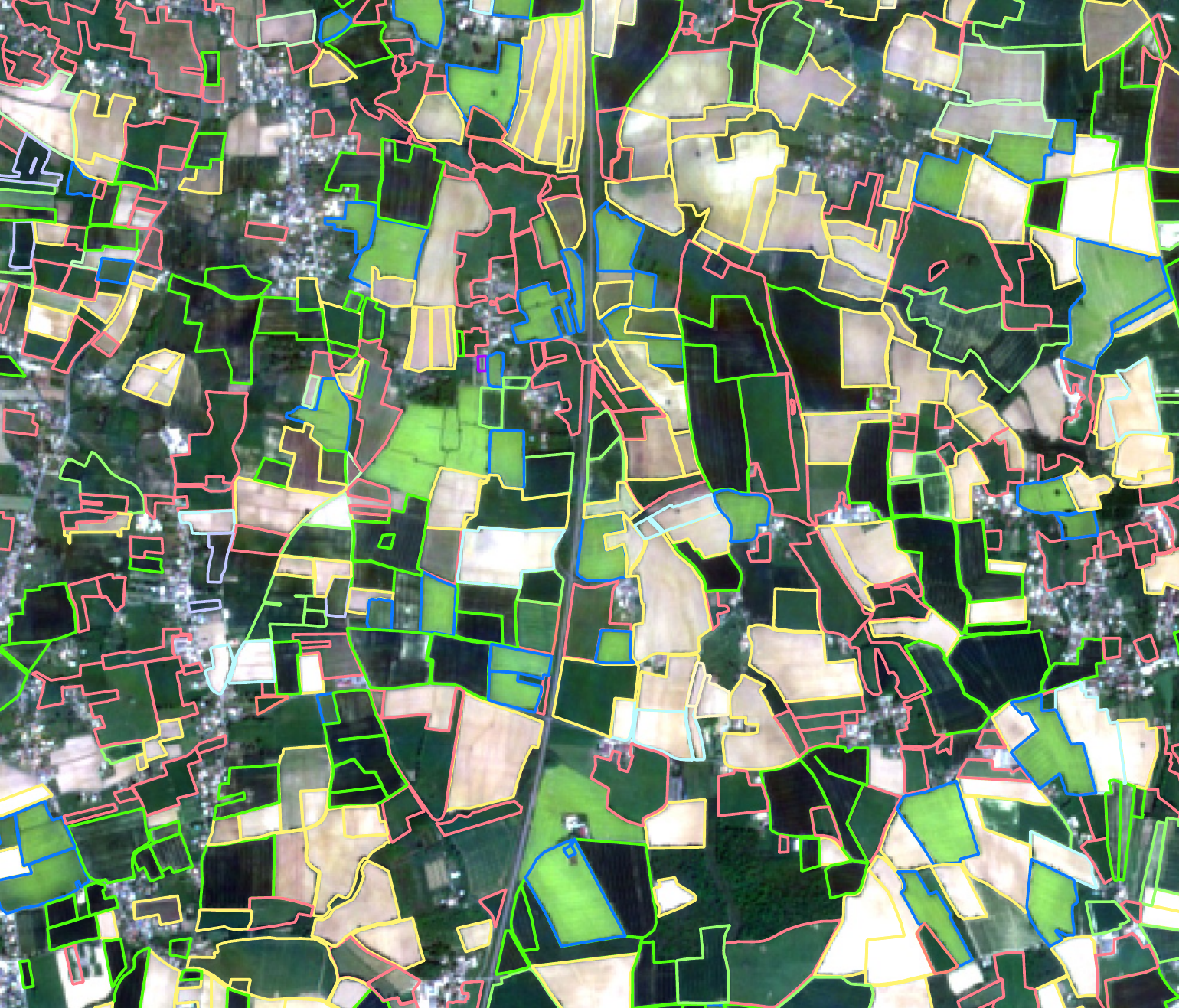}
    \end{subfigure}
    & 
    \begin{subfigure}{\ARXIV{0.31}{0.2}\textwidth}
        \centering\begin{tikzpicture}
    \node[anchor=south west,inner sep=0] (image) at (0,0) {\includegraphics[width=\textwidth]{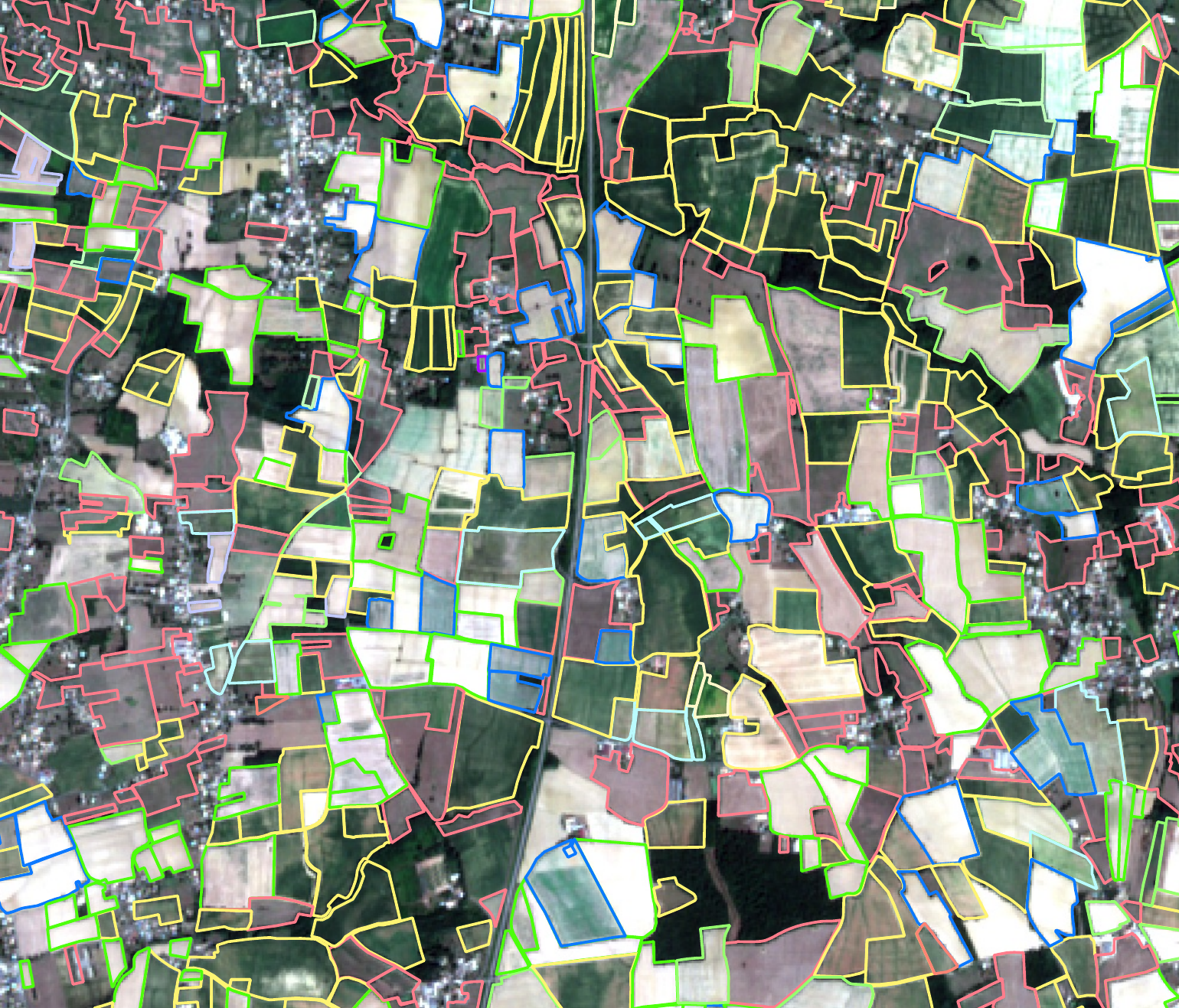}};
    \begin{scope}[x={(image.south east)},y={(image.north west)}]
        \draw[fill=none, draw=red] (0.5,0.65) rectangle (0.55,0.7);
        \node[fill=none, draw=none, text=black] (n1) at (0.9,0.9) {\includegraphics[width=\ARXIV{0.1}{0.1}\textwidth]{images/north.png}} ;
        \node[fill=none, draw=none, text=black] (n1) at (0.9,0.80) {\contour{white}{N}} ;
        \draw[fill=white, text=black, draw=none] (0.65,0.05) rectangle (0.95,0.13);
        \draw[<->, fill=white, text=black, draw=black] (0.65,0.10) -- (0.95,0.10);
        \draw[-, draw=black] (0.8,0.09) -- (0.8,0.11);
        \node[fill=none, text=black, draw=none] at (0.8,0.07)  {\tiny 1 km};
    \end{scope}
    \end{tikzpicture}
    \end{subfigure}\\
    \begin{subfigure}{\ARXIV{0.31}{0.1}\textwidth}
    \caption{}
    \label{fig:area_evolution:winter}
    \end{subfigure}
    &
    \begin{subfigure}{\ARXIV{0.31}{0.1}\textwidth}
    \caption{}
    \label{fig:area_evolution:spring}
    \end{subfigure}&
    \begin{subfigure}{\ARXIV{0.31}{0.1}\textwidth}
    \caption{}
    \label{fig:area_evolution:summer}
    \end{subfigure}
    \end{tabular}
    
    \caption{\textbf{Intra-Year Dynamics.} {Evolution of two areas across three seasons of the year $2020$. The top parcels contain mainly meadow parcels, while the bottom one comprises more diverse crops. The aspect of most parcel drastically changes across one year's worth of acquisition, corresponding to different phases in the growth cycle.} (\textbf{a}) Winter. (\textbf{b}) Spring. (\textbf{c}) Summer}
    \label{fig:area_evolution}
\end{figure}

{Our satellite time series are constituted of Sentinel-2 level 2A images. We discard the bands B01, B09, and B10 and resample the remaining $10$  spectral bands to a spatial resolution of $10$m per pixel with bilinear interpolation}. Our data spans three years of acquisition: $2018$, $2019$, and $2020$, with respectively $36$, $27$ and $29$ valid entries, see \figref{fig:area_evolution}. The length of sequences varies due to the automatic discarding of cloudy tiles by the data provider \href{https://www.theia-land.fr/en/product/sentinel-2-surface-reflectance/}{THEIA}  \cite{baghdadi2015theia}. We do not apply {further preprocessing such as cloud removal or radiometric calibration than what is already performed by the data provider \href{https://www.theia-land.fr/en/product/sentinel-2-surface-reflectance/}{THEIA}}.

We select \emph{stable} parcels, meaning that their contours only undergo minor changes across the three studied years. We also discard very small parcels (under $800$m$^2$) small or with very narrow shapes { to reflect the resolution of the Sentinel-2 satellite}. Each parcel has a ground truth cultivated crop type for each year {corresponding to the main culture as reported by the French LPIS}, whose precision is estimated at over $97$\% {as reported by} the French Payment Agency. {Note that we ignore secondary cultures for parcels with multiple growth cycles.} To limit class imbalance, we only keep crop types among a list of $20$ of the most cultivated species in the area of interest. In sum, our dataset is composed of $103\,602$ parcels, each associated with three image time sequences and three crop annotations {corresponding to the farmers' declarations for} $2018$, $2019$, and $2020$.

The Sentinel2Agri dataset \cite{garnot2020satellite}, composed of parcels from the same area, is composed of $191\,703$ parcels. We can estimate that our selection criteria exclude approximately every other parcel. A more detailed analysis of the evolving parcel partitions across different plots could lead to { retaining a higher proportion of the original parcels}.

As represented in \tabref{tab:class_breakdown}, the dataset is still imbalanced: more than $60$\% of {declarations} correspond to meadows. In comparison, potato is cultivated in less than $100$ parcels each year in the area of interest.

\begin{table}[ht!]
    \caption{\textbf{Crop distribution.} We indicate the number of parcels {declarations} in the LPIS for each class across all  $103\,602$ parcels and all $3$ years.}
    \label{tab:class_breakdown}
    \centering
    \small{\begin{tabular}{llll}
        \toprule
        Class & Count  & Class & Count \\
        \midrule
        Meadow & 184\,489 & Triticale & 5114\\
        Maize & 42\,006 & Rye & 569 \\
        Wheat & 27\,921 & Rapeseed & 7624\\
        Barley Winter & 10\,516 & Sunflower & 1886 \\
        Vineyard & 15\,461 & Soybean & 6072\\
        Sorghum & 820 & Alfalfa & 2682\\
        Oat Winter & 529 & Leguminous & 1454 \\
        Mixed cereal & 1061 & Flo./fru./veg. & 1079 \\
        Oat Summer & 330 & Potato & 230 \\
        Barley Summer & 538 & Wood pasture & 425\\
        \bottomrule
    \end{tabular}}
\end{table}

\subsection{Pixel-Set and Temporal Attention Encoders}
The Pixel Set Encoder (PSE) \cite{garnot2020satellite} is an efficient spatio-spectral encoder which learns expressive descriptors of the spectral distribution of the observations by randomly sampling pixels within a parcel. Its architecture is inspired by set-encoding deep architecture \cite{qi2017pointnet, kussul2017deep}, and dispenses us from preprocessing parcels into image patches, saving memory and computation. The Temporal Attention Encoder (TAE) \cite{garnot2020satellite} and its parsimonious version Lightweight-TAE (LTAE) \cite{garnot2020lightweight} are temporal sequence encoders based on the language processing literature \cite{vaswani2017attention} and adapted for processing SITS.
Both networks can be used sequentially to map the sequence of observations $x^i$ at year $i$ to a learned yearly spatio-temporal descriptor $e^i$:
\begin{align}
e^i = \TAE{\left[\PSE{x_t^i}\right]_{t=1}^{T_i}}~.
\end{align}
\begin{figure}[ht!]
    \centering
    \begin{tabular}{cc}
    \begin{subfigure}{\ARXIV{0.5}{0.35}\textwidth}
        \centering\includegraphics[width=1\textwidth]{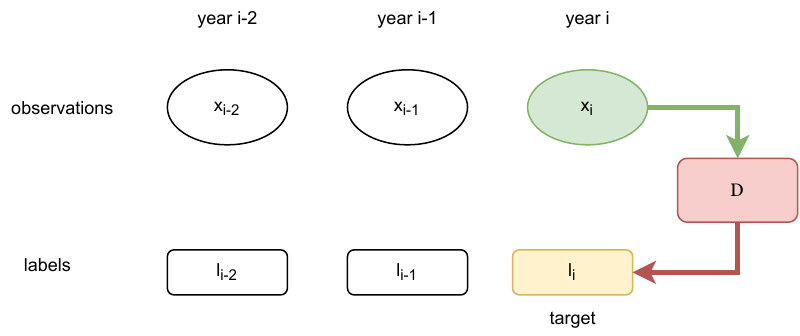}
    \end{subfigure}
         & 
    \begin{subfigure}{\ARXIV{0.45}{0.35}\textwidth}
        \centering\includegraphics[width=1\textwidth]{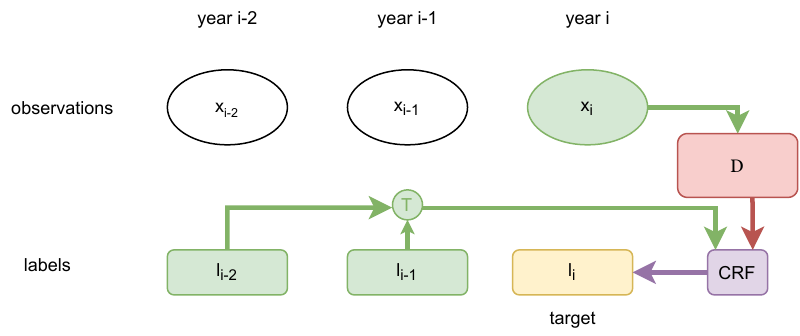}
    \end{subfigure} \\
    \begin{subfigure}{\ARXIV{0.45}{0.2}\textwidth}
    \caption{Single-year model\\}
    \label{fig:model:single}
    \end{subfigure}
    &
    \begin{subfigure}{\ARXIV{0.45}{0.2}\textwidth}
    \caption{CRF model\\}
    \label{fig:model:crf}
    \end{subfigure}\\
    \begin{subfigure}{\ARXIV{0.45}{0.35}\textwidth}
        \centering\includegraphics[width=1\textwidth]{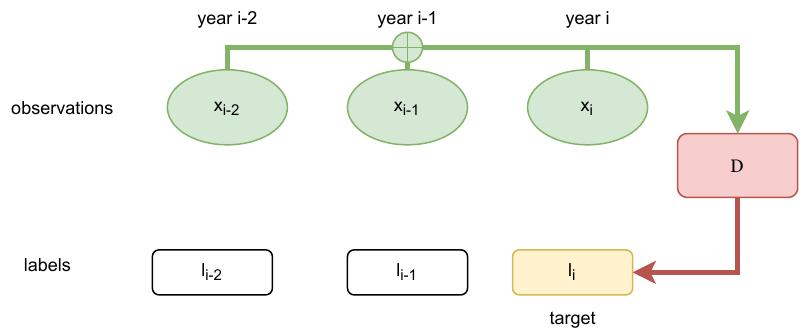}
    \end{subfigure}
         & 
    \begin{subfigure}{\ARXIV{0.45}{0.35}\textwidth}
        \centering\includegraphics[width=1\textwidth]{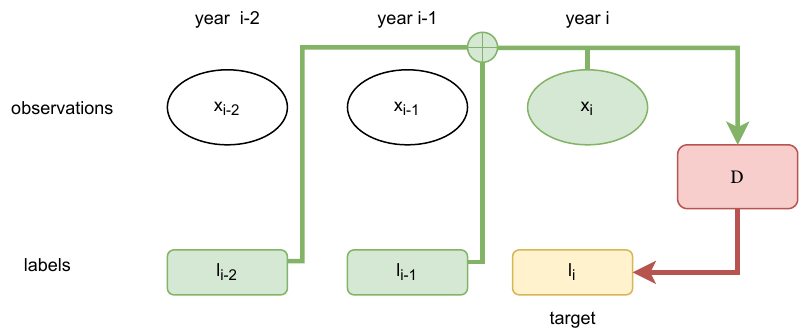}
    \end{subfigure} \\
    \begin{subfigure}{\ARXIV{0.45}{0.2}\textwidth}
    \caption{Observation bypass}
    \label{fig:model:obs}
    \end{subfigure}
    &
    \begin{subfigure}{\ARXIV{0.45}{0.2}\textwidth}
    \caption{Proposed model}
    \label{fig:model:dec}
    \end{subfigure}
    \end{tabular}
    \caption{\textbf{Multi-Year Modeling.} 
    Different approaches to model crop rotation dynamics: \Subref{fig:model:single} the model only has access to the current year's observation; 
    \Subref{fig:model:crf} a chain-CRF is used to model the influence of past cultivated crop;
    \Subref{fig:model:obs} the model has access to the observation of the past two years;
    \Subref{fig:model:dec} proposed approach: the model has access to the last two declared crops.}
    \label{fig:model}
\end{figure}
\subsection{Multi-Year Modeling}
\label{sec:dec}
We now present a simple modification of the PSE+LTAE network to model crop rotation.
In the original PSE+LTAE approach, the descriptor $e^i$ is directly mapped to a vector of class scores $z^i$ by a Multi-Layer Perceptron (MLP).
In order to make the prediction $z^i$ covariant with past cultivated crops, we augment the spatio-temporal descriptors $e^i$ by concatenating the sum of the one-hot-encoded labels $l^{j}$ for the previous two years $j=i-1, i-2$. Then, a classifier network $\mathcal{D}$, typically an MLP, maps this feature to a vector $z^i$ of $L$ class scores:
\begin{align}\label{eq:dec}
 z^{i} = \mathcal{D}\left(\left[ e^i \;\middle|\middle|\;  l_{i-1} + l_{i-2} \right]\right)~,
\end{align}
with $[\cdot||\cdot]$ the channelwise concatenation operator. 
We handle the edge effects of the first two available years by defining $l^0$ and $l^{-1}$ as vectors of zero of size $L$ (temporal zero-padding). This model can be trained end-to-end to simultaneously learn inter-annual crop rotations along with {the intra-annual evolution of the parcels' spectral statistics.}. 
Our model makes three simplifying assumptions:
\begin{itemize}
    \item We only consider the last two previous years because of the limited span of our available data. However, it would be straightforward to extend our approach to a longer duration.
    \item We consider that the history of a parcel is entirely described by its past cultivated crop types, and we do not take the past satellite observations into account. In other words, the label at year $i$ is independent from past observations conditionally to its past labels \cite[Chap~2]{wainwright2008graphical}.
    This design choice allows the model to stay tractable in terms of memory requirements.
    \item The labels of the past two years are summed and not concatenated. The order in which the crops were cultivated is then lost, but results in a more compact model.
\end{itemize}
\subsection{Baseline Models}
\label{sec:baselines}
In order to meaningfully evaluate the performance of our proposed approach, we implement different baselines {for classifying parcels from multi-year data} . In \figref{fig:model}, we represent schematically the main idea behind these baselines and our proposed approach. {Note that the choice of backbone network to handle single-year data is out of the scope of this paper.}\\

\textbf{Single-Year:}~$M_\text{single}$. We do not provide the previous years' labels, and directly map the current year's observations to a vector of class scores \cite{garnot2020lightweight}.\\

\textbf{Conditional Random Fields:}~$M_\text{CRF}$. Based on the work of \cite{bailly2018crop} and \cite{giordano2018temporal}, we implement a simple chain-CRF probabilistic model. We use the prediction of the previous PSE+LTAE, calibrated with the method of Guo \etal \cite{guo2017calibration} to approximate the posterior probability $p  \in [0,1]^L$ of a parcel having the label $k$ for year $i$ :  $p_k = P(l^i=k \mid x^i)$ (see \secref{sec:calibration} for more details). We then model the second order transition probability $p(l^i=k \mid l^{i-1}, l^{i-2})$ with a three-dimensional tensor $T \in[0,1]^{L \times L \times L}$ which can be approximated based on the observed transitions in the training set. As suggested by Bailly \etal, we use a Laplace regularization \cite[Chap.  13]{schutze2008introduction} to increase robustness. The resulting probability for a given year $i$ is given by:
\begin{align}
 {z}^{i}_\text{CRF}[k] = p \odot  T[l^{i-2}, l^{i-1}, :]~,
\end{align}
with $\odot$ the Hadamard term-wise multiplication.
This method is restricted to $i>2$ as edge effects are not straightforwardly fixed with padding.\\

\textbf{Observation Bypass:}~$M_\text{obs}$. Instead of concatenating the labels of previous years to the embedding $e^i$, we concatenate the average of the descriptors of the last two years  $e^{i-1}$ for $e^{i-2}$:
\begin{align}
 {z}^{i}_\text{obs} = {\mathcal{D}_\text{obs}}\left(\left[ e^i \;\middle|\middle|\;  
 \begin{cases}
 \frac12 [e^{i-1} + e^{i-2}] & \;\text{if}\;i>1 \\
 e^{0} & \;\text{if}\;i=1 \\
 0 & \;\text{if}\;i=0
 \end{cases}
 \right]\right).
\end{align}
Edge effects are handled  with mirror and zero temporal padding.\\

\textbf{Label Concatenation:}~$M_\text{dec-concat}$. Instead of concatenating the sum of the last two previous years, we propose to concatenate each one-hot-encoded vector $l^{i-1}$ and $l^{i-2}$ with the learned descriptor $z^i$. This approach is similar to \equaref{eq:dec}, but leads to a larger descriptor and a higher parameter count.
\\

\textbf{Single-Year Label Bypass:}~$M_\text{dec-one-year}$.
In order to evaluate the impact of describing the history of parcels as the past two cultivated crops, we only concatenate the label of the previous year to the learned descriptor $e^i$. 
\subsection{Training Protocol}
\begin{figure}[h]
\def\imagetop#1{\vtop{\null\hbox{#1}}}
    \centering
    \begin{tabular}{ccc}
    \imagetop{
    \begin{subfigure}{\ARXIV{0.4}{0.3}\textwidth}
        \centering\includegraphics[width=1\textwidth]{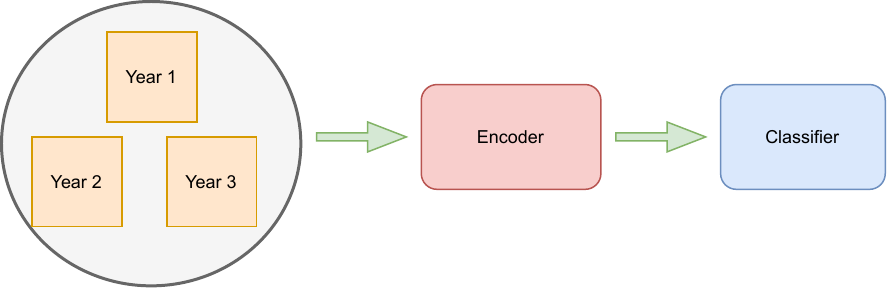}
    \end{subfigure}
    }
         & 
          \quad
         \imagetop{
    \begin{tikzpicture}
     \draw[thick, dotted, color=black!50!white] (0,2)-- (0,5);
    \end{tikzpicture}
    }
         \quad
         &
         \imagetop{
    \begin{subfigure}{\ARXIV{0.4}{0.3}\textwidth}
        \centering\includegraphics[width=1\textwidth]{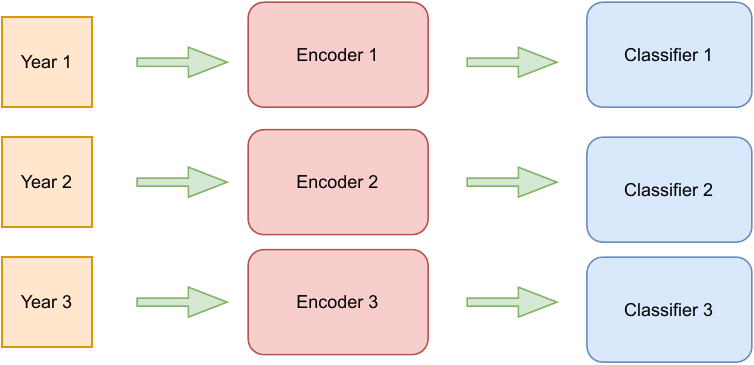}
    \end{subfigure}} \\
    \begin{subfigure}{\ARXIV{0.4}{0.2}\textwidth}
    \caption{Mixed-year training}
    \label{fig:mixed:mixed}
    \end{subfigure}
    &
    &
    \begin{subfigure}{\ARXIV{0.4}{0.2}\textwidth}
    \caption{Specialized models}
    \label{fig:mixed:spec}
    \end{subfigure}
    \end{tabular}
    \caption{\textbf{Training Protocol.} A single model is trained with parcels taken from all three years \Subref{fig:mixed:mixed}, and three specialized models whose training set only comprises observation for a given year \Subref{fig:mixed:spec}.}
    \label{fig:mixed}
\end{figure}
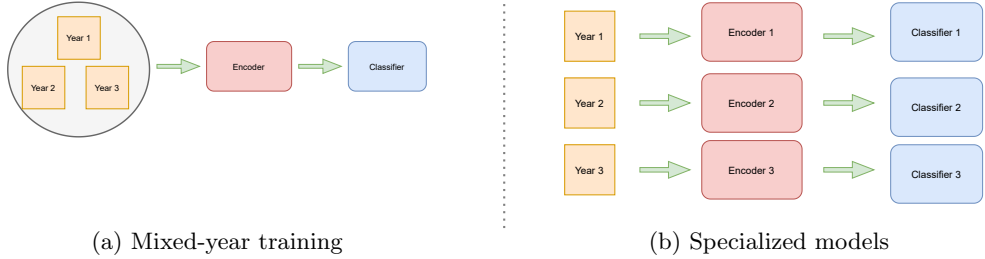
We propose a simple training protocol to leverage the availability of observations and {farmers' declarations} from multiple years.

\paragraph{ Mixed-year Training:} We train a single model with parcels from all available years. Our rationale is that exposing the model to data from several years will contribute to learning richer and more resilient descriptors. Indeed, each year has different meteorological conditions influencing the growth profiles of crops. {Moreover, by increasing the size of the dataset, mixed-year training mitigates the negative impact of rare classes on the performance.} 

We assess the impact of mixed-year training by considering $I=3$ \emph{specialized} models whose training set is restricted to a given year: $M_{2018}, M_{2019}$, and $M_{2020}$. In contrast, the model $M_\text{mixed}$ is trained with all parcels across all years with no information regarding of the year of acquisition. All models share the same PSE+LTAE configuration \cite{garnot2020lightweight}. We visualize the training protocols in \figref{fig:mixed}, and report the results in \tabref{tab:mixed}.

\paragraph{ Cross-validation:} We split our data into $5$ folds for cross-validation. For each fold, we train on $3$ folds and use the last fold for calibration and model selection{, corresponding to a train/validation/test ratio of $60\%,20\%$ and $20\%$ in each fold.} In order to avoid data contamination and self-correlation, our folds are all spatially separated: the fold separation is done parcel-wise and not for yearly observations. A parcel cannot appear in multiple folds for different years.



\subsection{Evaluation Metrics}
In order to assess the performance of the different approaches evaluated, we report the Overall Accuracy (OA), corresponding to the rate of correct prediction. {If we denote by $N_c$ the number of accurate predictions and $N$ the total number of parcels, the overall accuracy writes:}
\begin{align}
        \text{OA} = \frac{N_c}{N}~.
    \end{align}
To address the high class imbalance, we also report the mean Intersection over Union (mIoU), defined as the unweighted class-wise average of the intersection over Union (or Jaccard distance)  between the prediction and the ground truth for each class. {For a given class $i$, $\text{IoU}_i$ is defined as the ratio between the number of elements that are both predicted and labeled by class $i$ (the intersection, or true positives, and the number of elements that are either predicted or labeled as belonging to class $i$ (the union). In terms of binary classification (class $i$ \textit{vs.} not class $i$), this translates into the following formula: }
\begin{align}
    \text{IoU}_i = \frac{\text{TP}_i}{\text{TP}_i + \text{FP}_i + \text{FN}_i}~,
\end{align}
with $\text{TP}_i, \text{FP}_i$ and $\text{FN}_i$ the number of true positives,  false positives, and false negatives respectively.
\noindent
{The mIoU represents the average of the IoU calculated over the $K$ studied classes:}
\begin{align}
    \text{mIoU} = \frac1K \sum_{i=1}^{K}\text{IoU}_i~.
is\end{align}

\section{Results}

This section presents the quantitative and qualitative impact of our design choice in terms of training protocol and architecture.

\subsection{Training Protocol}

Predictably, the specialized models have good performance when evaluated on a test set composed of parcels from the year they were trained, and poor results for other years{, making this training procedure ill-fitted for the application at hand}. {In contrast, } the model $M_\text{mixed}$ vastly outperformed specialized models on average over the three considered years: over $15$ points of mIoU. More surprisingly, the model $M_\text{mixed}$ also outperforms all specialized models even when evaluated on the year of their training set. This implies that the increased diversity of the mixed-year training set allows the model to learn more robust and expressive representations.

\begin{figure}[p]
    \begin{tabular}{c}
    \begin{subfigure}{\ARXIV{.8}{0.8}\textwidth}
        \centering\includegraphics[width=\textwidth]{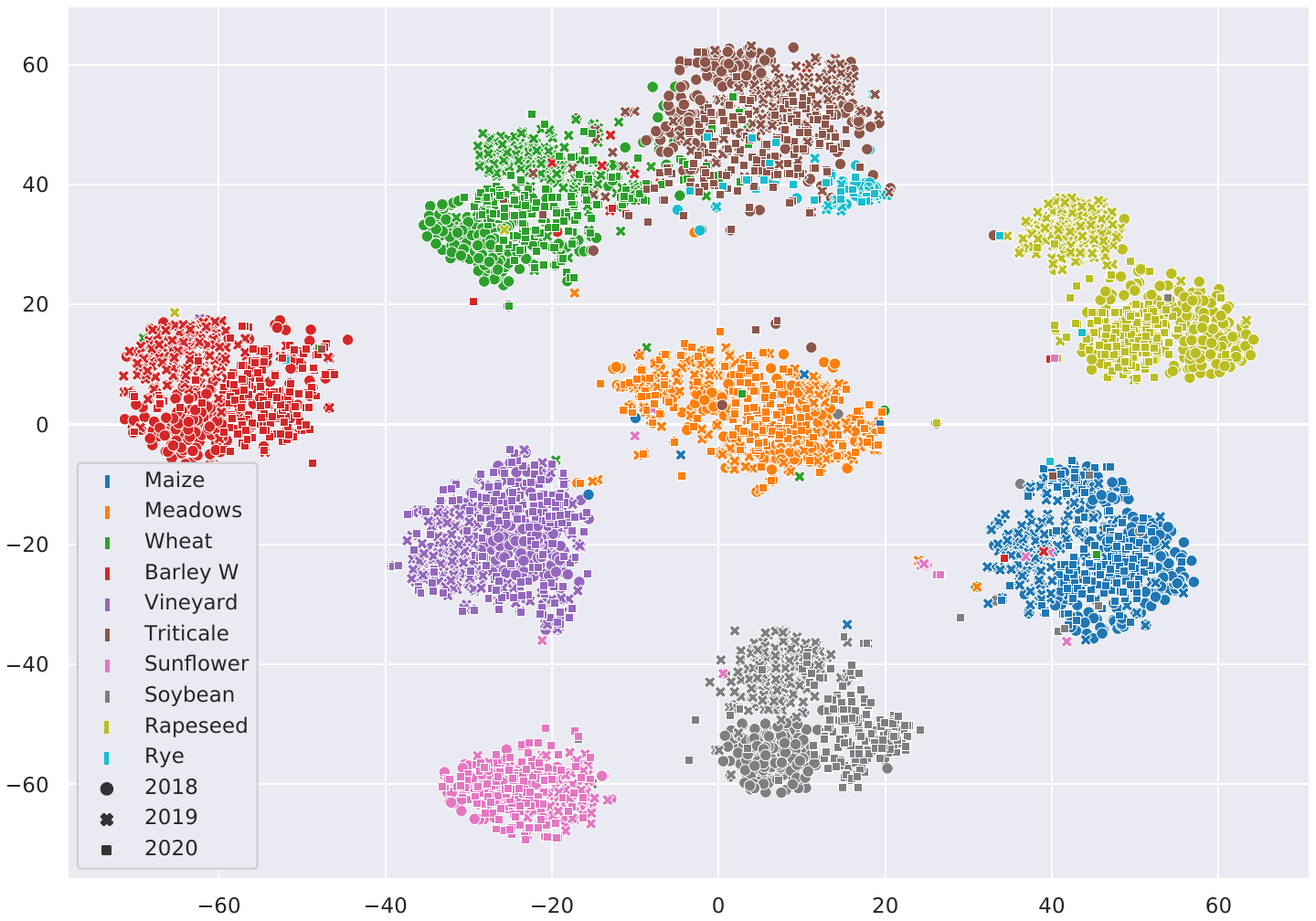}
    \end{subfigure}
    \\
    \begin{subfigure}{\ARXIV{.8}{0.08}\textwidth}
    \caption{$M_\text{mixed}$\\}
     \centering \label{fig:tsne:mixed}
    \end{subfigure}
    \\
    \begin{subfigure}{\ARXIV{.8}{0.8}\textwidth}
        \centering\includegraphics[width=\textwidth]{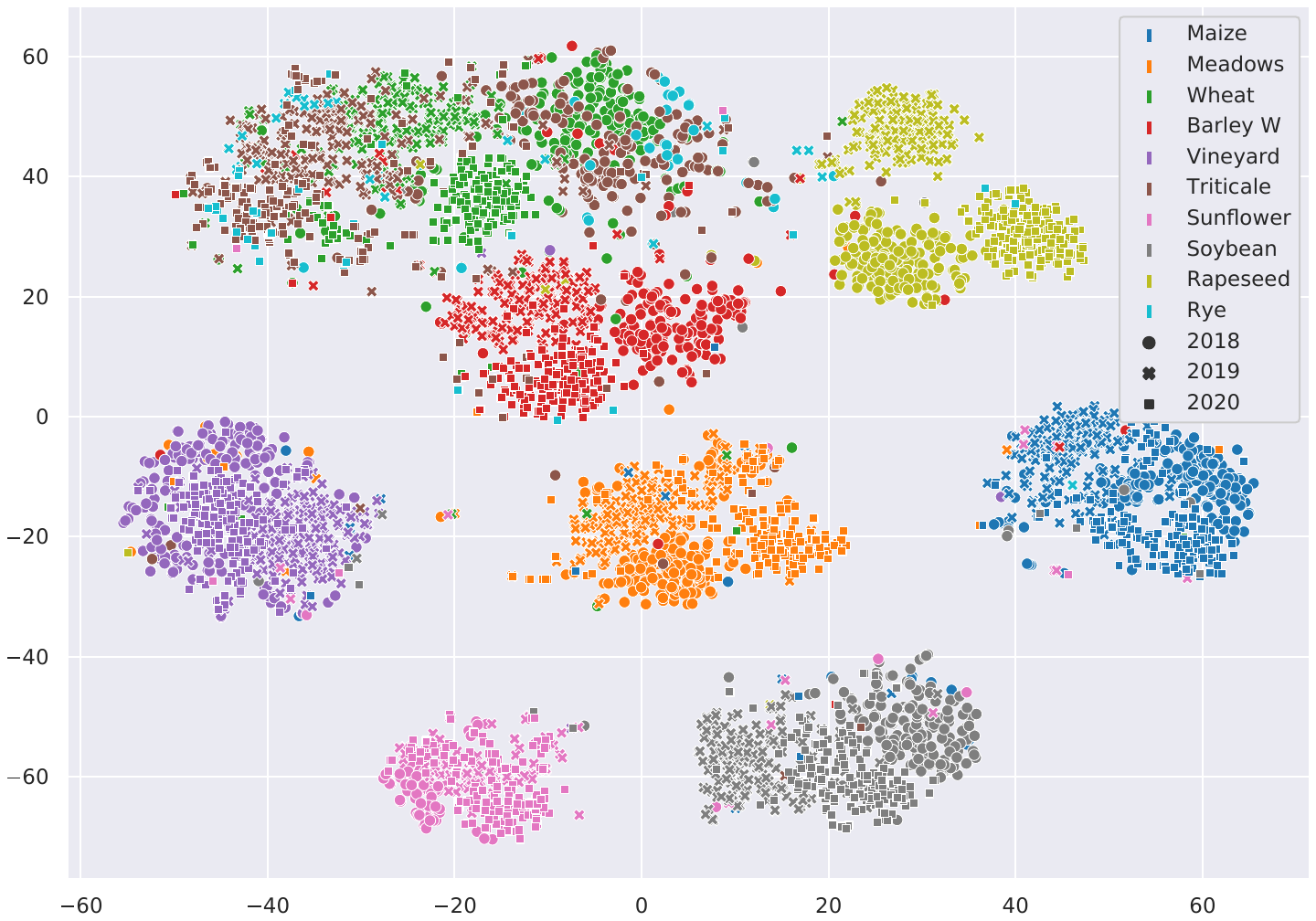}
    \end{subfigure}
    \\
    \begin{subfigure}{\ARXIV{.8}{0.08}\textwidth}
    \caption{$M_\text{2020}$}
    \label{fig:tsne:speci}
    \end{subfigure}
    \end{tabular}
    \caption{\textbf{Learned Representations.}  
   Illustrations of the learned SITS representations of the mixed-year model $M_\text{mixed}$ \Subref{fig:tsne:mixed} and the specialized $M_{2020}$ \Subref{fig:tsne:speci}. T-SNE algorithm is used to plot in $2$D the representation for $100$ parcels over $10$ classes and $3$ years.  We observe that $M_\text{mixed}$ produced cluster of embeddings that are consistent from one year to another, and with clearer demarcation between classes.
    }
    \label{fig:tsne}
\end{figure}

In \figref{fig:tsne}, we illustrate the representations learned by the mixed model $M_\text{mixed}$ and the specialized model $M_{2020}$. We remark that the parcel embeddings of the specialized model are inconsistent from one year to another, resulting in a higher overlap between classes. In contrast, the mixed year model $M_\text{mixed}$ learns year-consistent representations. This results in embedding clusters with large margins between classes, illustrating the ability of the model to learn robust and discriminative SITS embeddings. 


\begin{table}[h]
    \centering
    \caption{\textbf{Quantitative evaluation.}
    Performance (mIoU and OA) of the different specialized models $M_{2018}$, $M_{2019}$, $M_{2020}$ and of the mixed-years model $M_\text{mixed}$ evaluated on each year individually and all available years simultaneously with $5$-fold cross-validation. The best performances are shown in bold. Boxed values correspond to evaluations where the training and evaluation sets are drawn from the same year. The mixed-year model performs better for all years, even compared to specialized models.}
    In the rest of the paper, we use mixed year training for all models.
    \begin{tabular}{c c cc c cc c cc c cc}
        \multirow{2}{*}{Model} &&
        \multicolumn{2}{c}{2018} &&
        \multicolumn{2}{c}{2019} &&
        \multicolumn{2}{c}{2020} &&
        \multicolumn{2}{c}{3 years}\\\cline{3-4}\cline{6-7}\cline{9-10}\cline{12-13}
        && OA & mIoU && OA & mIoU && OA & mIoU && OA & mIoU \\
        \midrule\cline{3-4}
        $M_{2018}$ &&  \multicolumn{1}{|c}{97.0} &  \multicolumn{1}{c|}{64.7} && 90.3 & 45.5 && 90.8 & 43.4 &&92.7& 49.1 \\\cline{3-4}\cline{6-7}
        $M_{2019}$ && 88.9 & 39.5 &&  \multicolumn{1}{|c}{97.2} & \multicolumn{1}{c|}{70.1} &&88.7& 40.1 && 91.6& 48.0 \\\cline{6-7}\cline{9-10}
        $M_{2020}$ && 91.4 & 44.2 && 93.7 & 51.8 && \multicolumn{1}{|c}{96.7}& \multicolumn{1}{c|}{67.3} &&93.9& 54.0 \\\cline{9-10}\cline{12-13}
        $M_\text{mixed}$ && \bf 97.3& \bf 69.2 &&\bf97.4& \bf 72.2&&\bf96.8& \bf 68.7 &&  \multicolumn{1}{|c}{\bf 97.2}& \multicolumn{1}{c|}{\bf 70.4}\\\cline{12-13}
        \bottomrule
    \end{tabular}
    \label{tab:mixed}
\end{table}

\subsection{Influence of Crop Rotation Modeling}
\begin{figure}[h]
    \centering
    \begin{tabular}{cc}
        \begin{subfigure}{\ARXIV{.47}{0.35}\textwidth}
        \centering\includegraphics[width=\textwidth]{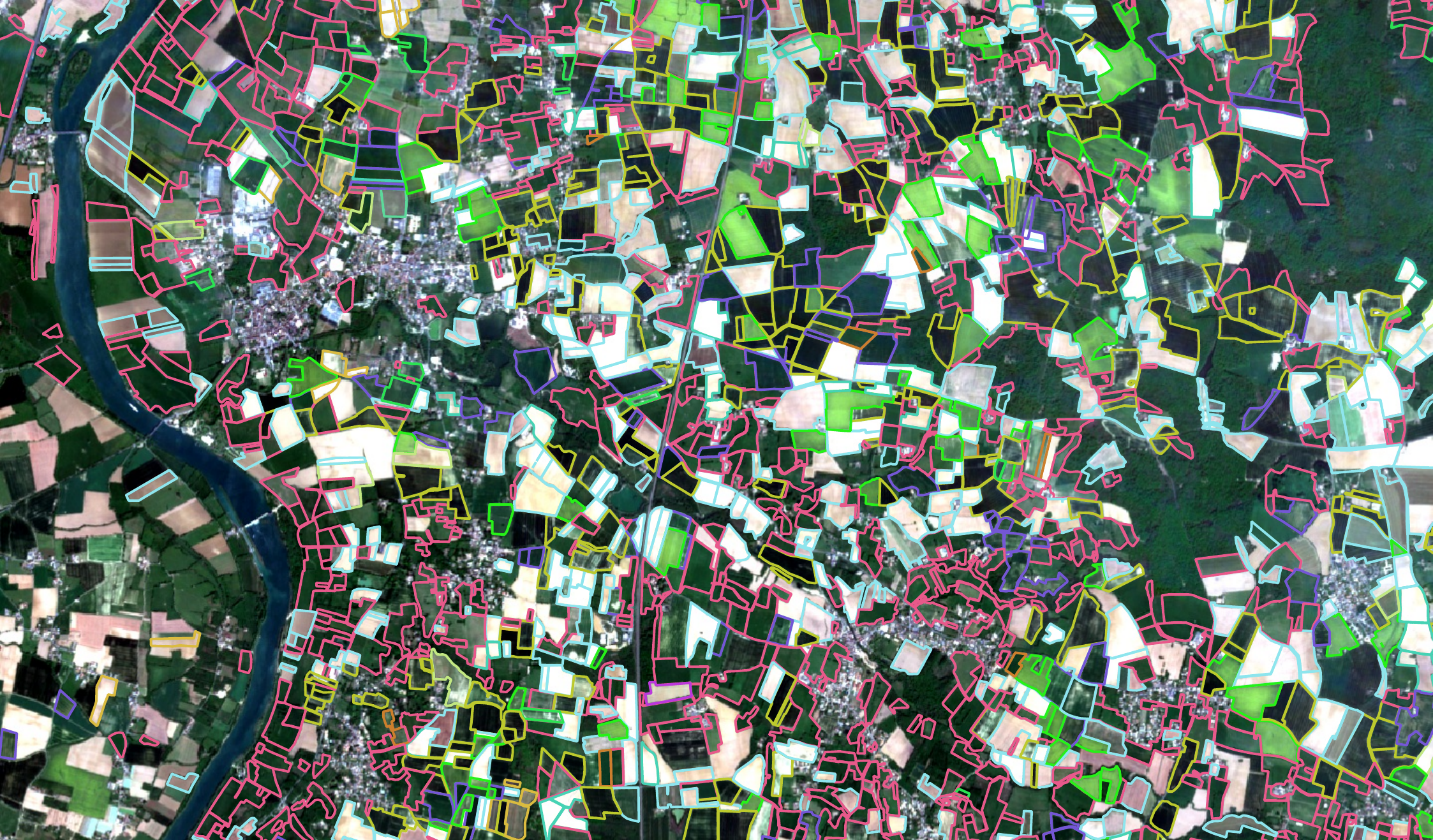}
    \end{subfigure}
    &
    \begin{subfigure}{\ARXIV{.47}{0.35}\textwidth}
        \centering
        \begin{tikzpicture}
    \node[anchor=south west,inner sep=0] (image) at (0,0) {\includegraphics[width=\textwidth]{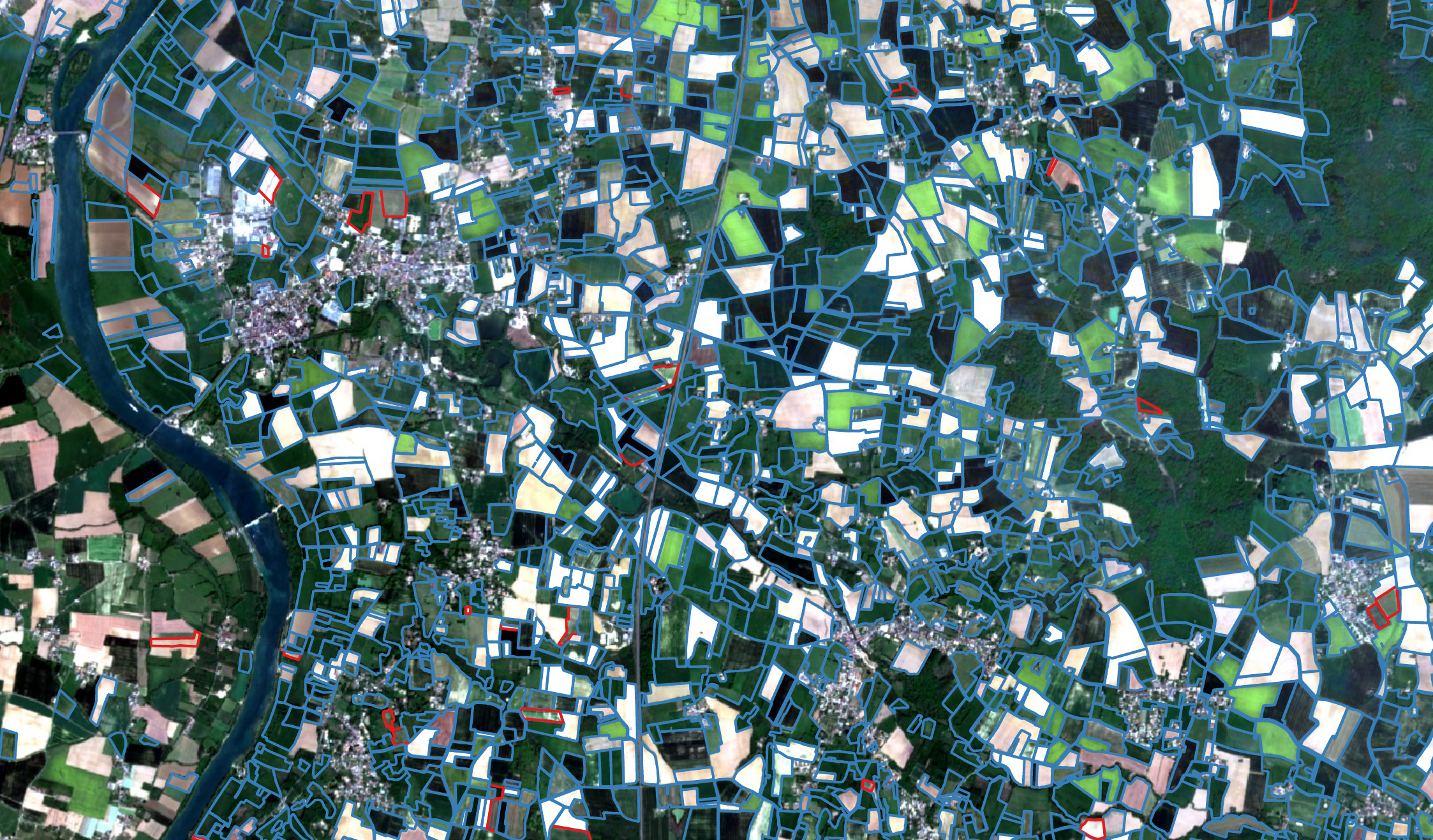}};
    \begin{scope}[x={(image.south east)},y={(image.north west)}]
        \draw[fill=none, draw=red] (0.5,0.65) rectangle (0.55,0.7);
        \node[fill=none, draw=none, text=black] (n1) at (0.9,0.9) {\includegraphics[width=\ARXIV{0.1}{0.1}\textwidth]{images/north.png}} ;
        \node[fill=none, draw=none, text=black] (n1) at (0.9,0.80) {\contour{white}{N}} ;
        \draw[fill=white, text=black, draw=none] (0.75,0.05) rectangle (0.95,0.13);
        \draw[<->, fill=white, text=black, draw=black] (0.75,0.10) -- (0.95,0.10);
        \draw[-, draw=black] (0.85,0.09) -- (0.85,0.11);
        \node[fill=none, text=black, draw=none] at (0.85,0.07)  {\tiny 2 km};
    \end{scope}
    \end{tikzpicture}

    \end{subfigure}
     \\
    \begin{subfigure}{\ARXIV{.47}{0.14}\textwidth}
    \caption{Ground truth}
    \label{fig:succes_by_parcel:gt}
    \end{subfigure}
    &
    \begin{subfigure}{\ARXIV{.47}{0.14}\textwidth}
    \caption{Errors}
    \label{fig:succes_by_parcel:success}
    \end{subfigure}
    \end{tabular}
    \caption{\textbf{Qualitative Illustration.} Detail of the area of interest with the ground truth in \Subref{fig:succes_by_parcel:gt} and the qualification of the prediction in \Subref{fig:succes_by_parcel:success} with correct prediction in blue and errors in red.}
    \label{fig:succes_by_parcel}
\end{figure}

We evaluate all models presented in \secref{sec:dec} and \secref{sec:baselines}, and provide qualitative illustration in \figref{fig:succes_by_parcel} .
All models are trained with the mixed-year training protocol and only tested on parcels from $2020$ to avoid edge effects affecting the evaluation. We give quantitative cross-validated results in \tabref{table:exp2}. Training our model on one fold takes $4$ hours, and inference on all parcels takes under 3 minutes (over $500$-parcels per second).

\begin{table}[h]
    \caption{\textbf{Performance by model.} Performances (mIoU and OA) of the models $M_\text{single}$, $M_\text{obs}$, $M_\text{CRF}$, and $M_\text{dec}$ tested for the year $2020$. Our proposed model $M_\text{dec}$ achieve higher performance than $M_\text{single}$ with a $6.3$\% mIoU gap. }
    \label{table:exp2}
    \centering
    \begin{tabular}{llll}
        \toprule
        Model & Description & OA & mIoU\\ 
        \midrule
        $M_\text{single}$ & single-year observation & 96.8 & 68.7 \\
        $M_\text{obs}$   & bypassing 2 years of observation & 96.8 & 69.3 \\
        $M_\text{CRF}$   & using past 2 declarations in a CRF & 96.8 & 72.3 \\
          $M_\text{dec-one-year}$   & concatenating last declaration only & 97.5 &  74.3 \\
        $M_\text{dec-concat}$   & concatenating past 2 declarations & 97.5 &  74.4 \\
        $M_\text{dec}$   & proposed method & \bf 97.5 & \bf 75.0 \\
        \bottomrule
    \end{tabular}
\end{table}

We observe that our model appreciably improved on the single-year model, with over $6$ points gained in mIoU. The CRF models also increase the results to a lesser margin. We attribute this inferior performance to an oversmoothing phenomenon already pointed out by Bailly \etal: CRFs tend to resolve ambiguities with the most frequent transition regardless of the specificity of the observation. In contrast, our approach simultaneously models the current year's observations and the influence of past cultivated crops.
$M_\text{obs}$ barely improves the quality of the single-year model. While this model has indeed access to more information than $M_\text{single}$, the same model is used to extract SITS descriptors for all three years. This means that the model's ambiguities and errors will be the same for all three representations, which prevent $M_\text{obs}$ from largely improving its prediction. Our approach injects new information into the model by concatenating the labels of previous years, which is independent of the model's limitations. Our method is more susceptible to the propagation of mistakes {in the farmers' declarations} but provides the most significant increase in performance in practice. 

Lastly, we concatenate both past label vectors to keep information about the order in which past crops were cultivated and observe a slight decrease in performance. The increase in model size can explain this. We conclude that this order is not  crucial information for our model conditionally to the observation of the target year.
Lastly, the model's performance with only the declaration of the last year performs almost as well as our model with two years worth of crop declarations. This suggests that yearly transition rules are sufficient to capture most inter-year dynamics, such as permanent culture. Alternatively, our two-year scheme may suffer from sharp edge effects with only three years' worth of data. Only a quantitative analysis over a longer period may resolve this ambiguity. {On average, our $M_\text{dec}$ model obtains an mIoU of $84.7$\% and overall accuracy of $98.1$\% on the training set.}

\begin{figure}[h]
\includegraphics[width=\ARXIV{.98}{0.7}\textwidth]{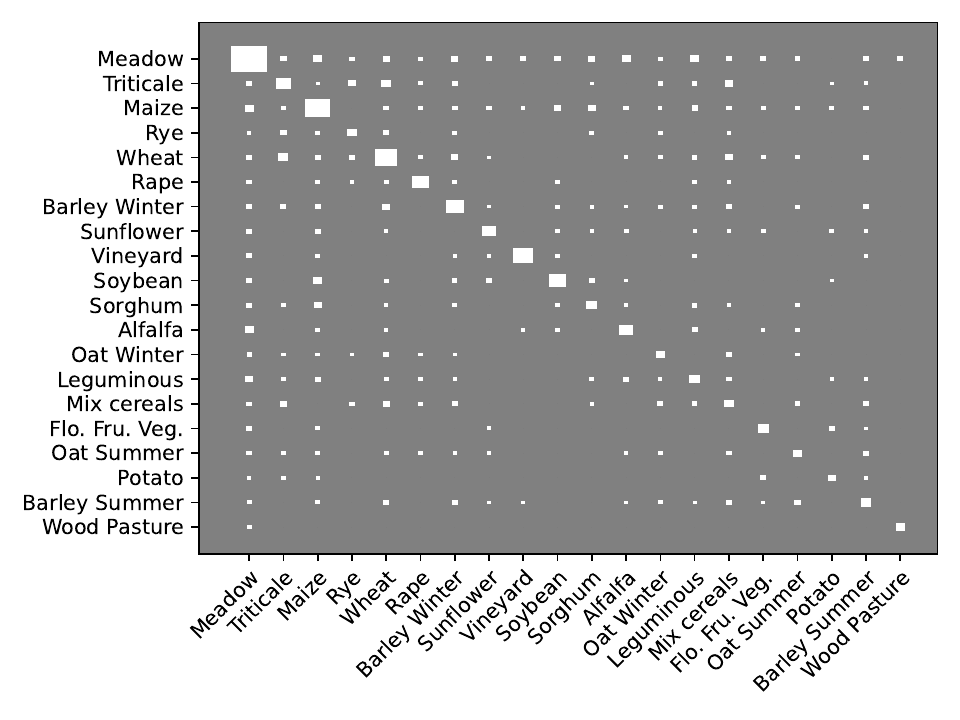}
\caption{\textbf{Confusion Matrix.} Confusion matrix of the prediction of $\mathcal{M}_\text{dec}$ for the year $2020$. The area of each entry corresponds to the square root of the number of predictions.}
\label{fig:hinton}
\end{figure}

We report the confusion matrix of $M_\text{dec}$  in \figref{fig:hinton}, and its performance for each crop in \tabref{tab:classwise}. We also compute $\Delta=\text{IoU}(M_\text{dec})-\text{IoU}(M_\text{single})$ the gain compared to the single-year model $\text{IoU}(M_\text{single})$, as well as the ratio of improvement $\rho=\Delta/(1-\text{mIoU}(M_\text{single}))$. This last number indicates the proportion of IoU that have been gained by modeling crop rotations. We observe that our model provides a large performance increase across all classes but four. The improvement is particularly stark for classes with strong temporal stability such as vineyards.

To further this analysis, we arrange the crop types into three groups according to the crop grown in $2018$ and the number of observed class successions over the $2018-2019-2020$ period:
\begin{itemize}
    \item {\bf Permanent Culture.} Classes within this group are such that at least $90\%$ of the observed successions are constant over three years. It Contains Meadow, Vineyard, and Wood Pasture.
    \item {\bf Structured Culture.} A crop is said to be structured if, when grown in $2018$, over $75\%$ of the observed three-year successions fall into $10$ different rotations or less and is not permanent. It contains Rapeseed, Sunflower, Soybean, Alfalfa, Leguminous, Flowers/Fruits/vegetables, and Potato.
    \item {\bf Other.} All other classes.
\end{itemize}
We report the unweighted class average for these three groups in \tabref{tab:groups}. Predictably, our approach considerably improves the results for permanent cultures. Our model is also able to learn non-trivial rotations as the improvement for structured classes is also noticeable. On average, our method also improves the performance for other nonstructured classes, albeit to a lesser degree. This indicates that our model can learn multi-year patterns not easily captured by simple rotation statistics.

\begin{table}[h]
    \caption{\textbf{Performance by class.} IoU per class of our model $M_\text{dec}$ for the year $2020$, as well as the improvement $\Delta$ compared to the single-year model $M_\text{single}$, and the ratio of improvement $\rho$. All values are given in $\%$, and we sort the classes according to decreasing ratios $\rho$.}
    \centering
    \small{\begin{tabular}{llllllll}
        \toprule
        Class & IoU & $\Delta$ & $\rho$ & Class &  IoU & $\Delta$ & $\rho$ \\\midrule
        Wood Pasture & 92.4 & +48.2 & 86.3 & Oat Summer & 52.8 & +3.6 & 7.0\\
        Vineyard & 99.3 & +1.4 & 68.7 &  Rapeseed & 98.3 & +0.1 & 6.6\\
        Alfalfa & 68.7 & +23.9 & 49.9 & Maize & 95.7 & +0.2 & 6.3\\
        Flo./Fru./Veg. & 83.4 & +14.5 & 46.5 & Wheat & 91.9 & +0.3 & 3.9\\
        Meadow & 98.4 & +0.9 & 36.9 & Barley Summer & 64.3 & +1.1 & 3.1 \\
        Leguminous & 45.2 & +14.6 & 21.1 & Potato & 57.1 & +0.5 & 1.2  \\
        Rye & 54.7 & +6.4 & 12.4 & Sunflower & 92.2 & -0.1 & -0.3\\
        Oat Winter & 57.7 & +4.5 & 9.7 & Sorghum & 56.6 & -0.2 & -0.4\\
        Triticale & 68.7 & +2.6 & 7.8 & Soybean & 91.8 & -0.2 & -3.1 \\
        Mix. Cereals & 31.0 & +5.1 & 6.8 & Barley Winter & 92.8 & -0.6 & -8.5\\
        \bottomrule
    \end{tabular}}
    \label{tab:classwise}
\end{table}

\begin{table}[ht!]
    \caption{\textbf{Improvement Relative to Structure.} Classwise IoU and mean improvement of our model compared to the single-year model according to the rotation structure of the cultivated crops.}
    \centering
    
    \small{\begin{tabular}{lll}
        \toprule
        Category & mIoU & mean $\Delta$\\
        \midrule
        Permanent & 97.3 & 16.9\\
        Structured & 77.7 & 7.6\\
        Other & 66.6 & 2.3\\
        \bottomrule
    \end{tabular}}
    \label{tab:groups}
\end{table}
\subsection{Model Calibration}
\label{sec:calibration}
Crop mapping can be used for various downstream applications, such as environmental monitoring, subsidy allocation, and price prediction. These applications carry crucial economic and ecological stakes and hence benefit from properly \emph{calibrated} prediction. A prediction is said to be calibrated when the confidence (\ie the probability associated to a given class) of the prediction corresponds to the empirical rate of correct prediction: we want $90\%$ of the prediction with a $90$\% confidence to be accurate. This allows for more precise risk estimation and improves control on the rate of false positives /  negatives.

\begin{figure}[h]
    \centering
    \begin{tabular}{cc}
    \begin{subfigure}{\ARXIV{0.47}{0.35}\textwidth}
        \centering\includegraphics[width=1\textwidth]{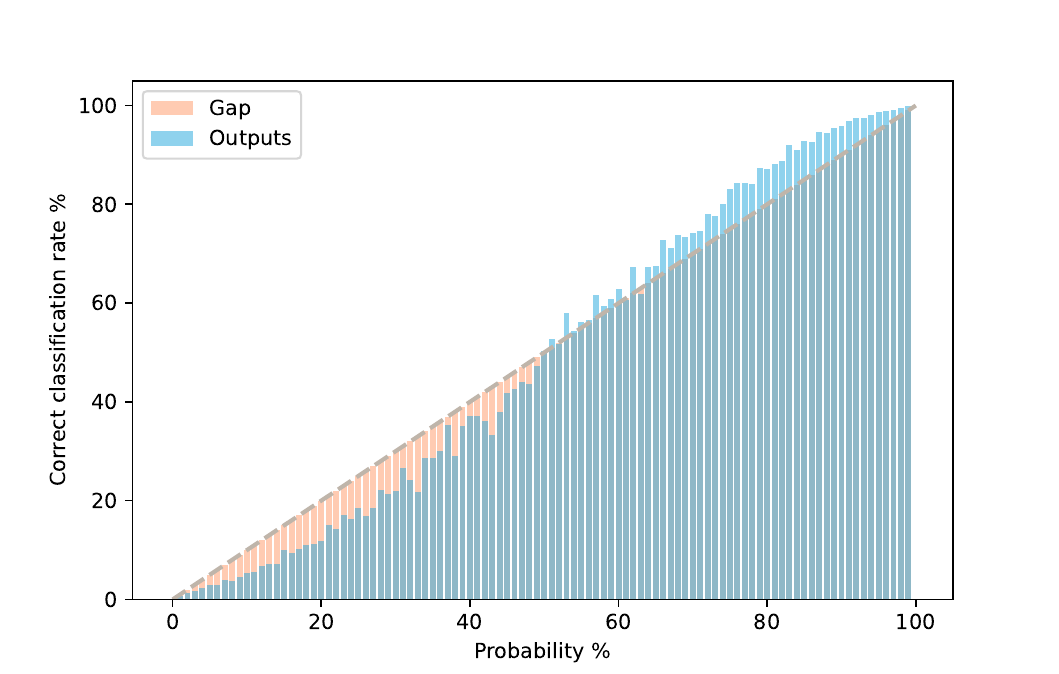}
    \end{subfigure}
         & 
    \begin{subfigure}{\ARXIV{0.47}{0.35}\textwidth}
        \centering\includegraphics[width=1\textwidth]{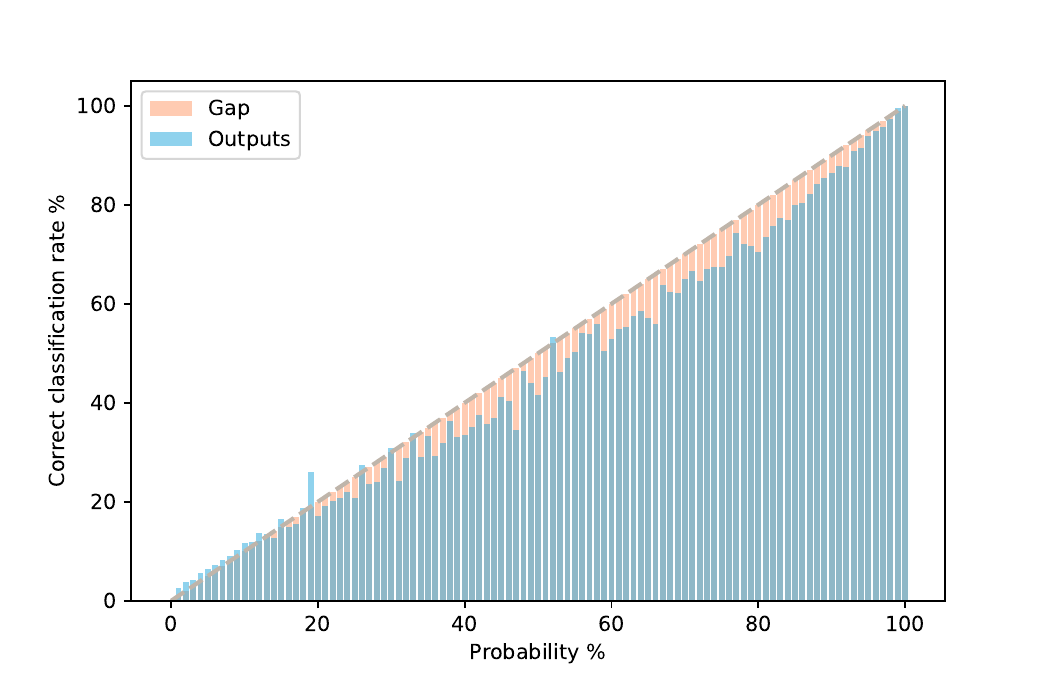}
    \end{subfigure} \\
    \begin{subfigure}{\ARXIV{0.47}{0.25}\textwidth}
    \caption{No calibration, ECE=1.4\%}
    \label{fig:calibration:before}
    \end{subfigure}
    &
    \begin{subfigure}{\ARXIV{.47}{0.25}\textwidth}
    \caption{Calibration, ECE=0.8\%}
    \label{fig:calibration:after}
    \end{subfigure}
    \end{tabular}
    \caption{\textbf{Model calibration.} 
    {Empirical rate of correct prediction by predicted confidence. We quantize the predicted confidence into $100$ bins for visualization purposes.} For a perfectly calibrated prediction, the blue histogram would exactly follow the orange line.  We observe that a simple post-processing step can considerably improve calibration.}
    \label{fig:calibration}
\end{figure}

Deep learning methods such as ours are notoriously badly calibrated. However, this can be corrected with the simple technique proposed by Guo~\etal  \cite{guo2017calibration}. {This method, called temperature scaling, consists in minimizing the discrepancy between the predicted confidence and the rate of errors binned into quantiles (we chose here $15$ bins) on the validation set by adjusting the temperature parameters in the last softmax layer \cite[Chap. 2.4]{bishop2006pattern}. } As represented in \figref{fig:calibration}, we can improve the calibration and observe a $43$\% decrease of the Expected Calibration Error (ECE) at a small computation cost.

\section{Discussion}

{In this paper, we set out to develop a deep learning method to leverage both the inter- and intra-annual dynamics of crop growth for crop mapping. 
We propose to enrich the learned spatio-temporal features with the last two declared cultures. Our experiments show that this simple method leads to an appreciable increase in performance compared to models operating on data drawn from  a single year. Our method outperforms other approaches such as CRF smoothing or observation stacking. This improvement can be observed for most crop types, including those with {rotation patterns beyond permanent culture}. We now discuss the limitations of our method and of our analysis.}

{\paragraph{Choice of Backbone Network. } Our method can be adapted to any network with a distinct classifier module mapping a spatio-temporal learned feature vector to a predicted vector of class scores. However, the choice of spatio-temporal encoder (\emph{backbone}) is out of the scope of this article. While it may be relevant to explore the effect of our modification on other architectures, we limited our investigation to the PSE+LTAE as it is the current state-of-the-art network for crop type mapping by a large margin.}

{\paragraph{Operational Setting.}
We showed that training our model with samples from all available years leads to considerably improved results. However, this scenario is not compatible with the operational setting of crop monitoring, in which payment agencies may want to detect erroneous declarations before all farmers' declarations have been received. Instead, we use the same setting as the vast majority of work in parcel classification and whose task is to classify parcels after the year is over \cite{zheng2015support,vuolo2018much,siachalou2015hidden,belgiu2018sentinel,kussul2017deep,russwurm2020self,pelletier2019deep,garnot2019time,russwurm2018multi,yuan2020self,kondmann2021denethor, garnot2020lightweight, garnot2020satellite, schneider2020re, garnot2021panoptic}.}

{As the Sentinel-2 mission was only operational starting in $2017$, we only have access to full-year coverage since $2018$. This means that we only have three years' worth of data at the time of writing this paper. In our opinion, this prevents us from a realistic setting in which the last year is withheld from the training set. Indeed, the inter-year meteorological variations between two years are typically too great to test for a third year and reasonably expect good results, as corroborated with preliminary experiments not shown in this paper. As more Sentinel-2 data become available, we will be able to evaluate our approach in a more realistic setting.}
 
{\paragraph{Scope of the Study.} Given the large amount of data involved and the complexity of data collection, we have limited our analysis and our proposed open-access dataset to a single area of the French Metropolitan territory. While nothing in our method is specific to this area, some of our analysis may be biased by the preponderance of stable cultures such as vineyards in this area. To confirm the generality of our conclusions, we would require a dataset with parcels taken from regions across the world with various meteorological conditions and agricultural practices. This task is complicated by the lack of harmonization between LPIS regarding open-access policy and even nomenclature. We hope that our results will encourage mapping agencies worldwide to release multi-year LPIS in open-source to help constitute a truly global dataset, allowing the community to assess the spatial generalizability of state-of-the-art methods.
We also limit ourselves to predicting the main culture in each parcel while ignoring cases with multiple growth cycles within one year. This may be particularly detrimental to its application in subtropical regions.}

{\paragraph{Applicability of our Model.} 
By requiring the last two grown crops to classify a parcel, our method cannot be applied to areas where the LPIS is not easily accessible. 
Furthermore, our training setting requires only selecting stable parcels. This can be easily obtained from the LPIS if it also contains information about the extent and position of each parcel, as is the case for the French LPIS. As a consequence, the effect of parcels with changing contours is out of the scope of our investigation.} 

\section{Conclusions}

We explored the impact of using multi-year data to improve the quality of the automatic classification of parcels from satellite image time series. 
We showed that training a deep learning model from multi-year observations improved its ability  to generalize and better precision across the board.
We proposed a simple modification to a state-of-the-art network to model both inter- and intra-year dynamics. This resulted in an increase of $+6.3\%$ of mIoU {when compared to models operating on single-year data.}
The effect is most substantial for classes with strong  temporal structures, but also impacts other crop types.
We also showed how simple post-processing could improve the calibration of the models considered.

Finally, we release both our code and our data. We hope that our promising results will encourage the SITS community to develop methods modeling multiple time scales simultaneously and  release more datasets spanning several years.

\FloatBarrier
\balance
\printbibliography

\end{document}